\documentclass[
reprint,
superscriptaddress,
preprintnumbers,
amsmath,
amssymb,
prc
]{revtex4-1}

\usepackage[     bookmarks,
                 bookmarksopen = true,
                 bookmarksnumbered = true,
                 linktocpage,
                 colorlinks = true,
                 linkcolor = blue,
                 urlcolor  = blue,
                 citecolor = blue,
                 anchorcolor = green,
                 hyperindex = true,
                 hyperfigures]
        {hyperref}
\usepackage{booktabs}
\usepackage{array}          
\usepackage{graphicx}
\usepackage{dcolumn}
\usepackage{bm}
\usepackage{float}
\usepackage{amsmath}
\usepackage{siunitx}

\usepackage{placeins} 
\usepackage{afterpage}

 \usepackage [utf8]{inputenc}
\usepackage[T1]{fontenc} 

\usepackage{subcaption} 
\usepackage{caption}
\usepackage{amsmath}

\usepackage{url} 
\usepackage{color}
\usepackage{graphicx}
\definecolor{dgreen}{cmyk}{1.,0.,1.,0.2}        
\definecolor{orange}{cmyk}{0.,0.353,1.,0.}

\usepackage{booktabs}
\usepackage[table]{xcolor}
\usepackage{colortbl}
\usepackage{graphicx}
\definecolor{BestBlue}{RGB}{31, 99, 214}   
\definecolor{SecondBlue}{RGB}{230, 240, 255}

\usepackage{soul}
\usepackage{tabularx}
\usepackage{ragged2e}
\usepackage{threeparttable}
\usepackage{makecell}
\usepackage{tabularray}
\UseTblrLibrary{booktabs}
\usepackage{adjustbox}

\definecolor{mincolor}{rgb}{0.8,0.85,1}
\sethlcolor{mincolor}

\definecolor{lightbrown}{rgb}{0.95,0.87,0.73}

\def \a {\alpha}

\newcommand\sect[1]{\section{#1}}

\begin{document}
\title{Physics-Informed Neural Network with Squeeze-Excitation-like Attention}

\author{Yun-Fei Song}
\affiliation{Key Laboratory of Quark and Lepton Physics (MOE) \& Institute of Particle Physics, Central China Normal University, Wuhan 430079, China}
\affiliation{Artificial Intelligence and Computational Physics Research Center, Central China Normal University, Wuhan 430079, China}

\author{Long-Gang Pang}
\email[]{lgpang@ccnu.edu.cn}
\affiliation{Key Laboratory of Quark and Lepton Physics (MOE) \& Institute of Particle Physics, Central China Normal University, Wuhan 430079, China}
\affiliation{Artificial Intelligence and Computational Physics Research Center, Central China Normal University, Wuhan 430079, China}

\author{Fu-Peng Li}
\email[]{fpli@fudan.edu.cn}
\affiliation{Key Laboratory of Quark and Lepton Physics (MOE) \& Institute of Particle Physics, Central China Normal University, Wuhan 430079, China}
\affiliation{Artificial Intelligence and Computational Physics Research Center, Central China Normal University, Wuhan 430079, China}
\affiliation{Key Laboratory of Nuclear Physics and Ion-beam Application (MOE) \& Institute of Modern Physics, Fudan University, Shanghai 200433, China}
\affiliation{Shanghai Research Center for Theoretical Nuclear Physics,
NSFC and Fudan University, Shanghai 200438, China}

\author{Jun-Jie Zhang}
\email[]{zjacob@mail.ustc.edu.cn}
\affiliation{Northwest Institute of Nuclear Technology, XiAn, 710024, China}

\date{\today}


\begin{abstract}

We introduce SEA-PINN, a novel architecture that incorporates a Squeeze-Excitation-like attention mechanism into physics-informed neural networks to dynamically recalibrate the importance of neurons across layers. A key feature of SEA-PINN is its highly stable initialization. On 17 out of 20 benchmark problems, SEA-PINN exhibit nearly negligible variance and significantly reduced initial loss, establishing a quasi-deterministic and favorable starting point for optimization. Notably, without employing Fourier feature embeddings or periodic activation functions, SEA-PINN attained competitive accuracy (83\% vs. 90\% improvement relative to FNN-PINN on the high-frequency case 7) as compared with TSA-PINN—a model specifically engineered for high-frequency problems via learnable frequencies in sinusoidal activations. Furthermore, integrating SEA-PINN into TSA-PINN boosted performance by 42.49\%. These results underscore SEA-PINN as a lightweight plug-in module that enhances nonlinear representation power, promotes more robust and efficient convergence, and strengthens the overall reliability of physics-informed learning.

\end{abstract}

\maketitle

\sect{Introduction}

Physics-Informed Neural Networks (PINNs) have emerged as a powerful approach for solving physics-based problems. These networks leverage the strong representational capability of deep neural networks and the computational efficiency of automatic differentiation \cite{Karniadakis2021,baydin2018automatic}. The concept built on early neural network applications to differential equations \cite{Lee1990,meade1994solution,dissanayake1994neural,Lagaris1998}, but gained prominence after Raissi et al. advanced it within modern deep learning frameworks \cite{hinton2006reducing, RAISSI2019}. PINNs integrate partial differential equations (PDEs) and boundary conditions into the loss function, enabling neural networks to learn the underlying physics and generate accurate solutions \cite{lu2021deepxde}. Their mesh-free property allows PINNs to handle complex geometries and irregular boundaries with ease, eliminating the need for predefined meshes and enabling sampling at arbitrary points in the solution domain. It is of great significance in high-dimensional domains to avoid the discretization errors inherent in spatial meshing by using PINNs, which represent solutions as continuous neural network functions. 
Additionally, PINNs employ automatic differentiation to compute PDE residuals, enabling self-supervised learning without the need for extensive and expensive datasets. This approach allows PINNs to generate physically consistent solutions in data-sparse or data-free environments. Leveraging these advantages, PINNs have been successfully applied to a broad spectrum of scientific and engineering disciplines, encompassing fluid mechanics \cite{jin2021nsfnets,qiu2022physics}, solid mechanics \cite{wang2022cenn,wu2024pinn}, metamaterial design \cite{chen2020physics,fang2019deep}, and seismic wave modeling \cite{ren2024seismicnet}.

The PINNs require retraining for each new set of PDE parameters, initial conditions, or boundary conditions. To address their limited generalization capability and computational inefficiency, novel architectures such as the Deep Operator Network (DeepONet) \cite{lu2021learning,wang2021learning,lu2022comprehensive,jin2022mionet,zhu2023fourier,jiang2024fourier,zhu2023reliable} and the Fourier Neural Operator \cite{lu2022comprehensive,li2021fourier} have been developed. These methods are designed to represent solution operators for a broad class of PDEs, significantly reducing data dependency \cite{jiao2025one}. However, operator learning demands a huge amount of training data, which are the paired (initial condition, boundary condition, PDE parameters) and the evolution in solution space from high-fidelity physical simulations.

The applications of PINNs in AI for Science is hindered by critical stability issues: solutions for identical problems exhibit high variance across multiple runs, with some runs failing to converge entirely, while performance degrades significantly for problems involving high-frequency solutions.
 Conventional PINNs prefer to learn low-frequency components first, while struggling to capture high-frequency, multi-scale, or sharp-gradient features \cite{Rahaman2019}. This is named as the spectral bias problem inherent from fully connected networks. To counter this, Fourier feature embeddings are used to map the solution into the frequency space, thereby enabling the network to capture high frequency components more effectively \cite{Tancik2020}. 
To further improve the representation capability, various attempts have been used to adjust the network structures used in PINNs.
For instance, employing locally adaptive activation functions \cite{jagtap2020locally} or integrating specialized network architectures, such as Convolutional Neural Networks (CNNs) \cite{gao2021phygeonet} for grid-structured data, Graph Neural Networks (GNNs) \cite{pfaff2020learning} for unstructured meshes, and Transformers for capturing long-range dependencies \cite{zhao2023pinnsformer}, into the PINNs framework. The domain-decomposition like Conservative PINNs (CPINNs) \cite{Jagtap2020} and Extended PINNs (XPINNs) \cite{Jagtap2021}, which enable parallelization and multi-physics coupling by training independent networks on subdomains.
Concurrently, adaptive sampling methods like Residual-Based Adaptive Refinement (RAR) focus computational resources on high-residual regions \cite{wu2023}.

Recent studies have identified several key factors contributing to potential training failures in PINNs and have proposed corresponding improvements. For time-dependent PDEs, standard full spatio-temporal training violates temporal causality, which can be addressed through sequential training strategies such as Causal PINNs \cite{Wang2022a}. Another significant issue is gradient imbalance resulting from the multi-component loss function, often indicative of an ill-conditioned optimization landscape; this has been mitigated via adaptive loss weighting \cite{wang2022and,mcclenny2023self} or gradient-statistics-based optimization \cite{Wang2021}. Furthermore, insufficient arithmetic precision (e.g., FP32) can lead to premature convergence due to numerical underflow; upgrading to FP64 has been demonstrated to resolve this failure mode and enable accurate solutions \cite{xu2025}.

Despite these advances, fundamental limitations persist: the convergence speed of PINNs is slow, and the performance across multiple runs is highly dependent on the initialization points in the high-dimensional parameter space. To address these issues, we propose incorporating a squeeze-excitation-like attention network (SEA-Net) into Physics-Informed Neural Networks (PINNs). The original Squeeze-and-Excitation Network (SENet) \cite{Hu2017}, designed for CNNs, dynamically assigns attention weights to different channels by computing these weights from the average pooling values of these channels and further adjusting their relative importance through a squeeze-excitation block.  
In our SEA-PINN, we adapt this idea to the neuron level: for each hidden layer, the complete set of neuron outputs is taken as input to a lightweight squeeze-excitation weight generator, which produces a neuron-wise weight vector. These weights rescale the neuron activations through element-wise multiplication before they are passed to the next layer. This design introduces a dynamic, input-dependent reweighting mechanism while keeping the backbone architecture of standard FNN-PINN unchanged.

The solution represented by our SEA-Net, $u_{\theta}(t, \vec{x})$, exhibits a notable characteristic: both the function values $u$ and their derivatives of various orders (e.g., $u_t$, $u_x$, $u_{xx}$, $\cdots$) have nearly zero mean and variance for sampled space-time coordinates across multiple runs with different random seeds. As a result, the residual loss for PDEs is very low even before training begins, as shown by:

\begin{align}
\delta = a_0 u + a_1 u_t + a_2 u_x + a_3 u_{xx} + \cdots \approx 0
\end{align}

This property contributes to the superior performance of SEA-PINN across a wide range of problems. For cases involving non-zero initial conditions, boundary conditions, or source terms in PDEs, our network performs at least as well as traditional approaches. A key advantage of SEA-PINN is the stability of solutions across multiple runs, along with rapid convergence in the early stages of training. Moreover, without the need for periodic activation functions or Fourier transformations, the network demonstrates strong performance in handling high-frequency problems. 

\sect{Results}
We evaluated SEA-PINN on the 20 solvable PDE problems for PINNs and compared the results with benchmark PINN model \cite{zhongkai2024pinnacle}.

\subsection{Lower Initial Loss and Fluctuations Drive Better Convergence}

\begin{figure*}[!ht]
\includegraphics[width=\linewidth,keepaspectratio]{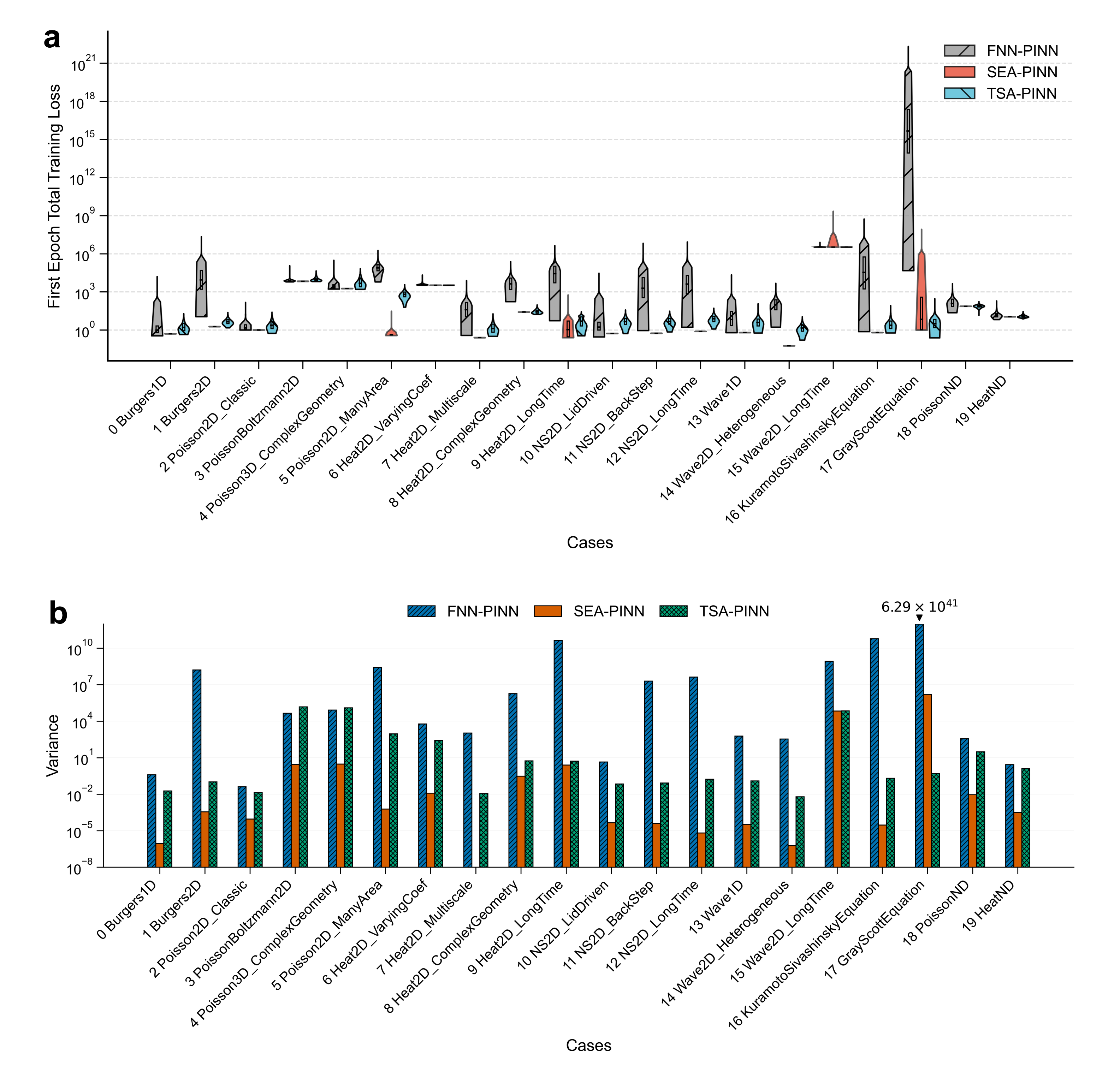} 
\caption{Early-training performance comparison of PINNs methods. \textbf{(a)} Distribution of the initial total training loss (first epoch) for FNN-PINN, SEA-PINN, and TSA-PINN across 20 benchmark cases, derived from 1000 independent runs per case (seeds 1-1000). \textbf{(b)} The bar charts present the loss variance at the tenth epoch from 1,000 random experiments.}
\label{fig:initial_and_10th_epoch}
\end{figure*}

Fig. \ref{fig:initial_and_10th_epoch}a presents the initial loss distribution of SEA-PINN in comparison with FNN-PINN across 1000 runs initialized with different random seeds, after trained for 1 epoch. The mean loss values of SEA-PINN are notably lower than those of FNN-PINN and TSA-PINN in 17 out of the 20 problems examined. Qualitatively, the loss variances of SEA-PINN are negligible relative to the other two methods. These results indicate that SEA-PINN not only improves training accuracy but also demonstrates significantly enhanced stability.

Here we focus on the total physics-informed training loss and its components, as this quantity is directly minimized during optimization and reveals how SEA-PINN balances PDE and boundary-condition terms at initialization and in the early phase of training. The final predictive accuracy will be assessed in Section~D using the relative $L^2$ error on test sets.

\begin{figure*}[htbp] 
\centering
\includegraphics[width= \linewidth,keepaspectratio]{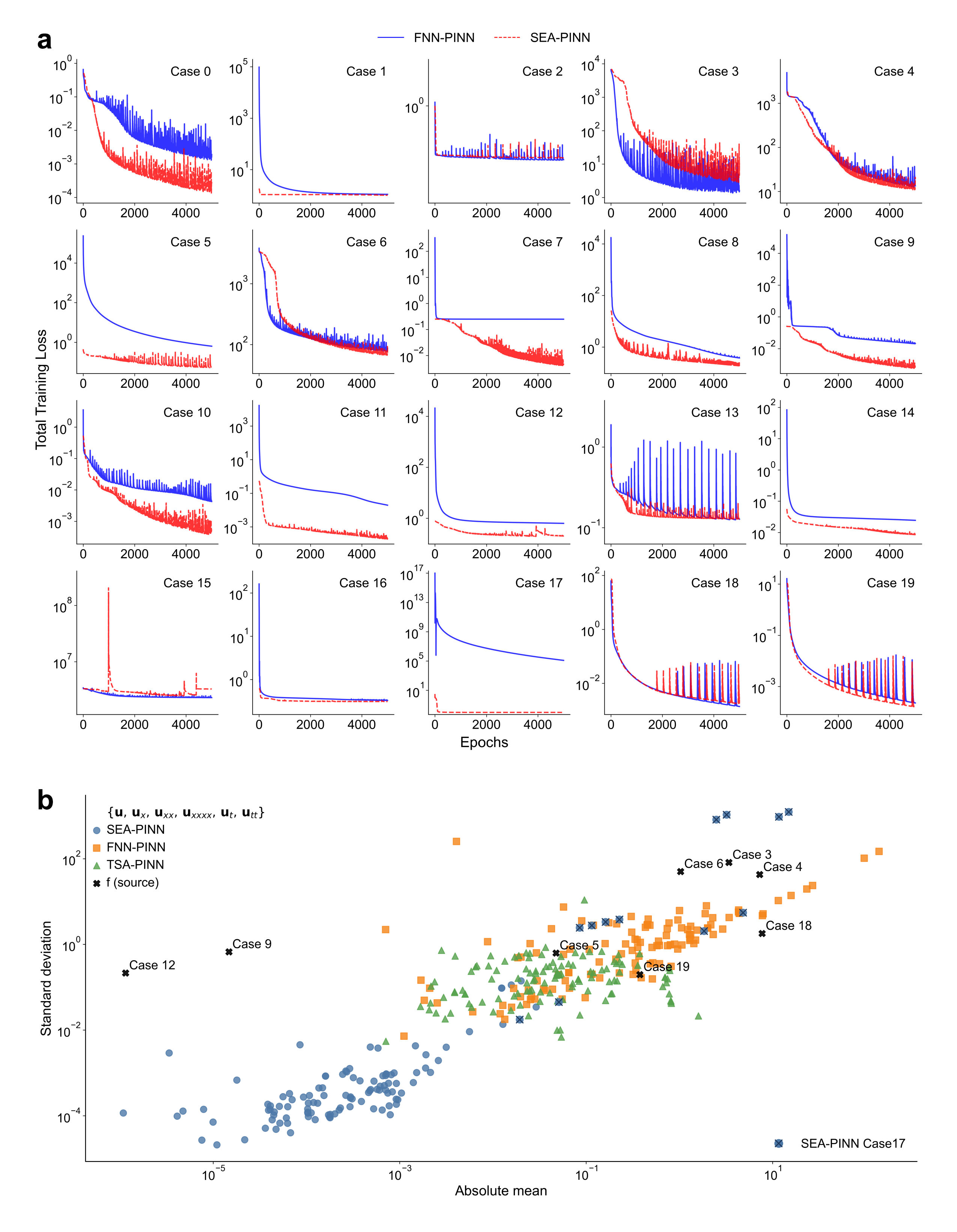} 

\caption{Difference from Ground Truth and training convergence. \textbf{(a)} Comparison of the total training loss between FNN-PINN and SEA-PINN over 5,000 epochs for 20 cases under a fixed random seed (seed=1). \textbf{(b)} Scatter plot comparing the means of absolute values and standard deviations for the predicted solution $u$, its derivatives, and the governing equation's source term $f$ (which depends only on sampling points) across all sampling points for FNN-PINN, SEA-PINN, and TSA-PINN in each case. The source term $f$ is network-independent.}
\label{fig:scatter_and_loss_comparison}
\end{figure*}

Fig. \ref{fig:initial_and_10th_epoch}b quantitatively compares the loss variance of multiple runs between our SEA-PINN and the other two methods at the 10th epoch, presented on a logarithmic scale. In all cases, the loss variance of our SEA-PINN is smaller than that of FNN-PINN, indicating superior stability in the initial state. The variance of SEA-PINN differs by several orders of magnitude from that of FNN-PINN, highlighting its significant advantage.
As an innovative architecture that enhances PINN performance through adjustable frequency mechanisms and dynamic slope recovery, Trainable Sine Activation PINN (TSA-PINN) exhibits lower initial loss and smaller variance compared to FNN-PINN. However, its distribution remains considerably wide, indicating that its stability improvement is substantially less pronounced than that of SEA-PINN. In 18 out of 20 cases, the loss variance of SEA-PINN is several orders of magnitude lower than that of TSA-PINN. This exceptional initialization performance, characterized by consistently lower initial loss and reduced variance, is a distinctive feature of our SEA-PINN network, providing a robust foundation for more effective and reliable training.

In traditional PINNs, a large initial loss—although it may drop rapidly—often constrains the attainable final loss and leads to suboptimal convergence. The initial loss of our SEA-PINN is very small, in many cases, even smaller than FNN-PINN  after trained for $5,000$ epochs. 

Fig. \ref{fig:scatter_and_loss_comparison}a compares the total training loss of SEA-PINN and FNN-PINN for a fixed seed (seed = 1) across all 20 PDE problems. The comparisons for other seeds (seed = 10, 20, 30) are provided in Supplementary Figs. 17--19. For the majority of cases, SEA-PINN starts from a much lower loss than FNN-PINN and maintains a clear advantage throughout training, converging to smaller final losses. 
Quantitatively, SEA-PINN achieves lower final loss than FNN-PINN in 15 out of the 20 cases, and is at least comparable in the remaining ones. 
This behavior indicates that SEA-PINN does not merely optimize faster, but also initializes in a more favorable region of the loss landscape. 

To understand why SEA-PINN starts from such a good initial state, we examine the statistics of the network outputs and their derivatives at initialization. 
For each case, we compute the means and standard deviations of the predicted solution $u$, its derivatives, and the source terms $f$ appearing in the governing equations on the same set of sampling points for FNN-PINN, SEA-PINN, and TSA-PINN. 
These quantities are summarized in the scatter plot of Fig.~\ref{fig:scatter_and_loss_comparison}b, where each marker corresponds to a particular equation term in one case.
Overall, the SEA-PINN points are concentrated near the lower-left corner, indicating smaller means and standard deviations than the other two networks. 
Because the PDE residual is a linear combination of $u$, its derivatives, and $f$, terms that are already close to zero in both mean and variance at initialization directly translate into smaller residuals and, consequently, a smaller initial PDE loss.  
This explains why SEA-PINN consistently starts from a low-loss regime and provides a more stable optimization trajectory. 

A few cases deviate from this general trend and help clarify the limitations of the initialization effect. 
In Cases 3, 4, and 6, the initial losses of both FNN-PINN and SEA-PINN are on the order of $10^{3}$.
Fig. \ref{fig:scatter_and_loss_comparison}b shows that these cases are associated with large means and standard deviations in the external source term $f$, which is independent of the network architecture. 
The large magnitude and variability of $f$ dominate the residual, amplifying the initial loss.
Case 18 also contains an external source term with a mean comparable to those of Cases 3, 4, and 15, but its standard deviation is two orders of magnitude smaller, leading to a more moderate initial loss on the order of $10^{2}$. 
In Case 17 the SEA-PINN points are shifted away from the lower-left corner, but the FNN-PINN points move even further towards the upper-right, resulting in substantially larger means and standard deviations and a much larger initial loss. 
These examples highlight that SEA-PINN's benefits are most pronounced when the governing equation does not contain numerically large, network-independent source terms. 

Across all benchmark problems, Supplementary Table 2 summarizes which loss components have zero target values and how they behave during the early training phase. 
We observe a consistent pattern: whenever a boundary or initial condition is identically zero, the corresponding loss term of SEA-PINN remains low from the beginning, whereas FNN-PINN typically requires a longer optimization process to reduce the same term. 

\begin{figure*}[htbp] 
\centering
\begin{tabular}{c}
    \includegraphics[width=1\linewidth,keepaspectratio]{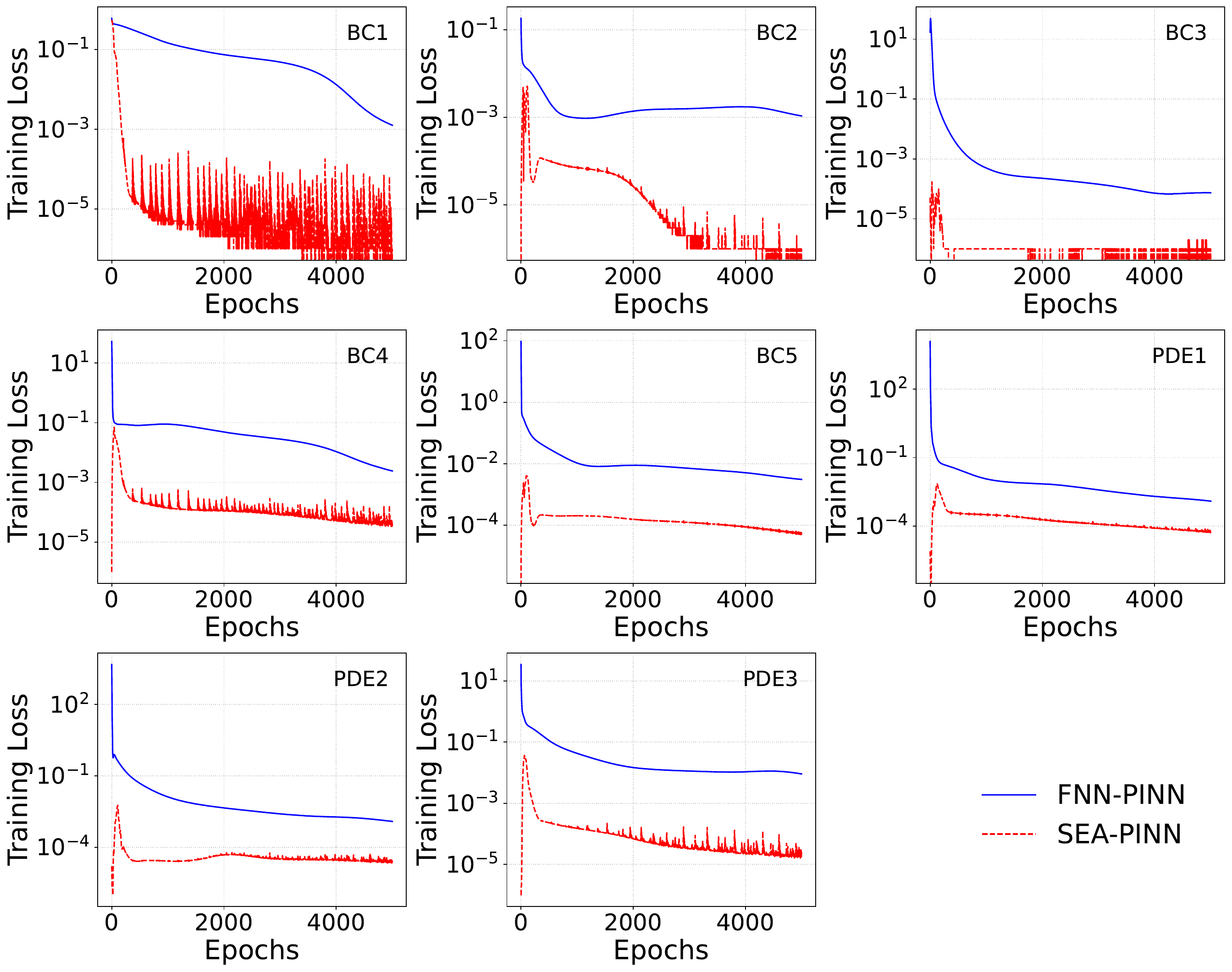}
\end{tabular}
\caption{Comparison of loss components for NS2D\_BackStep (Case 11)}
\label{fig:Comparison_of_loss_components_for_NS2D_BackStep(Case11)}
\end{figure*}

\subsection{Inherent Fulfillment of PDEs and BCs in SEA-PINN: Loss Dynamics and Balancing}

Fig. \ref{fig:Comparison_of_loss_components_for_NS2D_BackStep(Case11)} illustrates the training loss curves of eight components (3 PDE loss terms + 5 BC loss terms) for Case 11 (Navier-Stokes Flow) as functions of training epochs. In this case, the loss components corresponding to zero boundary conditions (BC2-BC5) and zero source terms (PDE1-PDE3) begin with minimal values at epoch 0, demonstrating that our SEA-PINN inherently satisfies three PDEs and four out of five boundary conditions without requiring any training. For the non-zero boundary condition BC1, the loss of SEA-PINN decreases significantly faster compared to FNN-PINN. Concurrently, the losses of other components exhibit a brief increase before subsequently decreasing, indicating a balancing mechanism between minimizing BC1 loss and other loss terms.

This scenario represents a typical multi-task optimization problem in PINNs with eight distinct training objectives. Traditional FNN-PINN initiates with substantially higher loss values across different components, exhibits slower convergence rates, and frequently displays pronounced oscillatory behavior during training. Furthermore, FNN-PINN encounters difficulties in automatically adjusting the relative weights among these loss terms to achieve minimal total loss. In contrast, since multiple objectives are inherently satisfied at initialization, our SEA-PINN begins with considerably fewer effective training targets, thereby facilitating more efficient optimization.

To demonstrate that the above characteristics consistently hold across all cases, Supplementary Table 1 summarizes the governing equations, domains, and the corresponding boundary/initial conditions for five representative benchmark cases. In addition to Case 11 (Fig.~\ref{fig:Comparison_of_loss_components_for_NS2D_BackStep(Case11)}), Supplementary Figs. 6--9 provide comparisons of all loss components for the other four cases. Moreover, Supplementary Figs. 1--5 report the total training loss  comparisons of the five cases over ten random seeds (seed = 1--10). Collectively, these supplementary results further corroborate that, by inherently satisfying multiple PDE/BC constraints at initialization, SEA-PINN achieves better convergence than FNN-PINN.

\subsection{Smaller output--input gradient norms and better inter-layer conditioning}

\begin{figure*}[htbp]
\centering
\includegraphics[width=\linewidth,keepaspectratio]{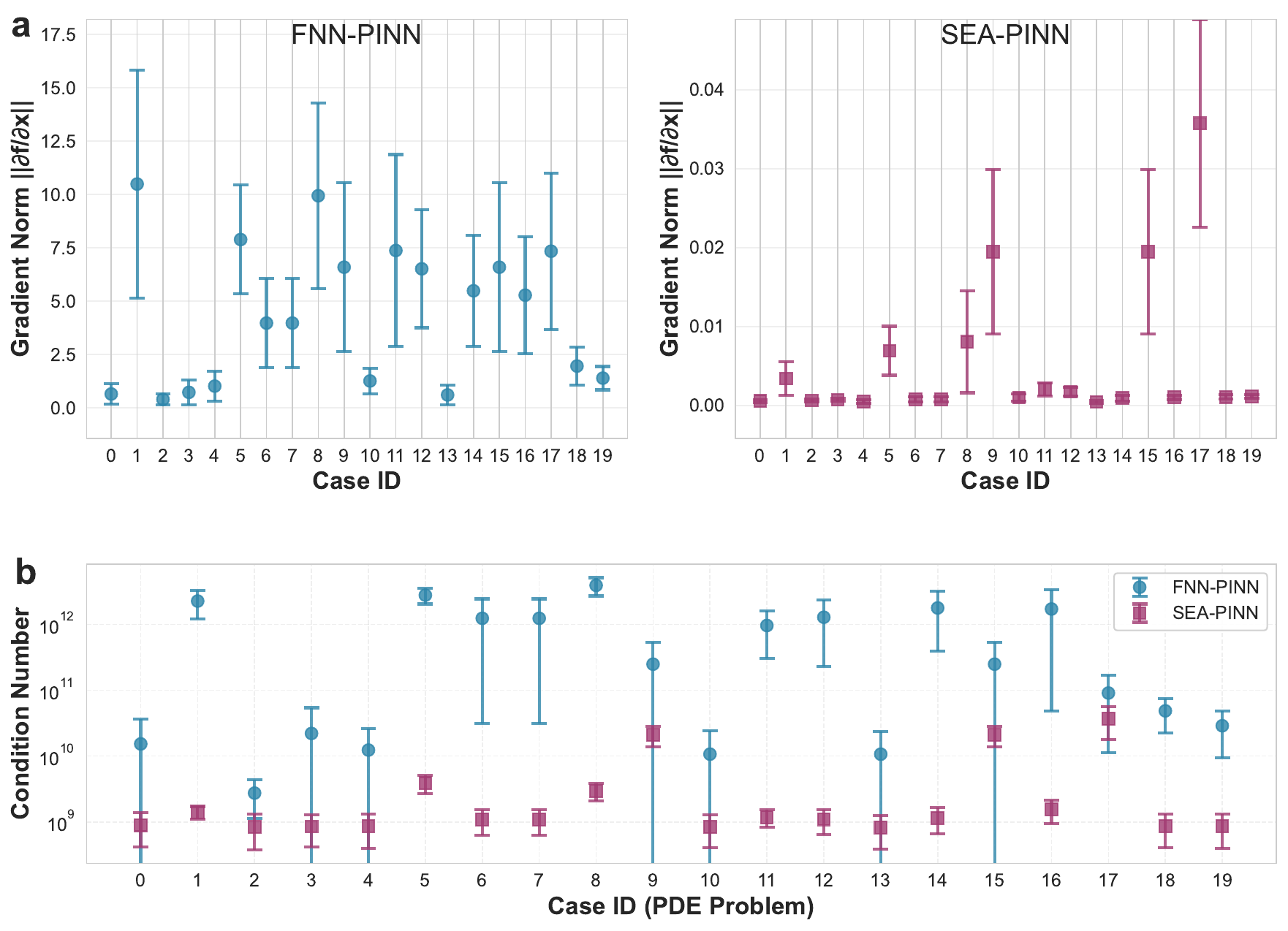}
\caption{Jacobian-based sensitivity of FNN-PINN and SEA-PINN at initialization.
\textbf{(a)} Cross-seed statistics of the output--input gradient norms $\|\partial \mathbf{f} / \partial \mathbf{x}\|_F$ at initialization across 20 benchmark PDE cases.
For each case and each of the five random seeds, the Jacobian norm is first averaged over $1,000$ randomly selected training points; the markers show, for each case, the mean of these per-seed averages and the error bars denote the corresponding standard deviation.
\textbf{(b)} Inter-layer condition numbers of the Jacobian mapping the output of the first hidden layer to that of the ninth hidden layer, computed on the same training points and five random seeds as in (a).
Markers show the mean condition number over five random seeds and error bars denote the corresponding standard deviation. 
}
\label{fig:gradient_norm_and_condition_number}
\end{figure*}

To further understand why SEA-PINN exhibits more stable initialization and improved convergence, we analyze the Jacobian of the network output with respect to the input, as well as the conditioning of deep feature transformations \cite{szegedy2013intriguing, schoenholz2016deep, pennington2017resurrecting, novak2018sensitivity, ross2018improving, pennington2018emergence}. Based on the Jacobian expressions derived in Methods (Section: Jacobian analysis of FNN-PINN and SEA-PINN), we first evaluate the Frobenius norm of the output--input Jacobian $\|\frac{\partial \mathbf{f}}{\partial \mathbf{x}}\|_F$ at initialization. 
For each of the 20 benchmark PDE problems and for each of the five random seeds, we randomly select $1,000$ training points from the actual collocation set and compute the corresponding output--input gradient norms. As shown in Fig.~\ref{fig:gradient_norm_and_condition_number}a, the error-bar plots show that SEA-PINN consistently achieves lower mean gradient norms with substantially smaller standard deviations than FNN-PINN across all cases. This indicates that SEA-PINN realizes a smoother and less sensitive input--output mapping at initialization, which is consistent with the smaller and less fluctuating initial losses reported in Figs. \ref{fig:initial_and_10th_epoch} and~\ref{fig:scatter_and_loss_comparison}.

We further investigate the conditioning of information propagation between hidden layers by measuring the condition number of the Jacobian that maps the output of the first hidden layer to that of the ninth hidden layer. For each case and each architecture, we again sample $1,000$ training points and compute the inter-layer Jacobian for five independent random seeds. The condition number for each seed is averaged over the $1,000$ samples, and we then aggregate the mean and standard deviation across the five seeds. Fig.~\ref{fig:gradient_norm_and_condition_number}b summarizes these results for all 20 PDE problems. SEA-PINN exhibits uniformly smaller mean condition numbers and smaller error bars than FNN-PINN, demonstrating that its deep feature transformations are better conditioned and less sensitive to perturbations. Together with the layer-wise Jacobian analysis (Supplementary Fig. 11), which shows that the Frobenius norms of the local Jacobians $\|\mathbf{D}^{(l)}\|_F$ are smaller for SEA-PINN than for FNN-PINN in representative cases (Cases~7 and~11), these results provide a mechanistic explanation for the reduced output--input gradient norms of SEA-PINN and its superior initialization stability: SEA-PINN simultaneously achieves smaller output--input gradient norms and improved inter-layer conditioning, rather than shrinking gradients at the cost of ill-conditioned Jacobians.

\subsection{More Stable and Lower Final Relative Error Across Multiple Experiments}

\begin{table*}[!t]
\centering
\caption{Performance Comparison of FNN-PINN and SEA-PINN Models on PDE Cases}
\label{tab:model_comparison}
\begin{tabular}{c l S[table-format=1.2e-1] S[table-format=1.2e-1] c}
\hline 
{Case ID} & {Case Name} & \multicolumn{1}{c}{FNN-PINN} & \multicolumn{1}{c}{SEA-PINN} & {Improvement} \\
& & {Rel. $L^2$ Error} & {Rel. $L^2$ Error} & (\%) \\
\hline 
0  & Burgers1D                  & 1.24e-1 &\textcolor{red}{\num{8.63e-2}} & \textbf{30.23} \\
1  & Burgers2D                  & 5.52e-1 & \textcolor{red}{\num{5.31e-1}} & \textbf{3.77}  \\
2  & Poisson2D\_Classic         & \textcolor{red}{\num{6.86e-1}} & 6.95e-1 & -1.25 \\
3  & PoissonBoltzmann2D         & 7.91e-1 & \textcolor{red}{\num{7.77e-1}} & \textbf{1.85}  \\
4  & Poisson3D\_ComplexGeometry & 7.00e-1 & \textcolor{red}{\num{5.60e-1}} & \textbf{19.92} \\
5  & Poisson2D\_ManyArea        & 9.87e-1 & \textcolor{red}{\num{6.20e-1}} & \textbf{37.19} \\
6  & Heat2D\_VaryingCoef        & \textcolor{red}{\num{2.54e0}}  & 3.74e0  & -47.23 \\
7  & Heat2D\_Multiscale         & 8.84e-1 & \textcolor{red}{\num{1.49e-1}} & \textbf{83.09} \\
8  & Heat2D\_ComplexGeometry    & 1.70e-1 & \textcolor{red}{\num{1.29e-1}} & \textbf{24.13} \\
9  & Heat2D\_LongTime           & 9.99e-1 & 9.99e-1 & -0.01 \\
10 & NS2D\_LidDriven            & 2.10e-1 & \textcolor{red}{\num{9.87e-2}} & \textbf{52.98} \\
11 & NS2D\_BackStep             & 5.72e-1 & \textcolor{red}{\num{1.09e-1}} & \textbf{80.90} \\
12 & NS2D\_LongTime             & 1.00e0  & \textcolor{red}{\num{9.97e-1}} & \textbf{0.28}  \\
13 & Wave1D                     & 5.87e-1 & \textcolor{red}{\num{5.67e-1}} & \textbf{3.28} \\
14 & Wave2D\_Heterogeneous      & \textcolor{red}{\num{1.35e0}}  & 1.52e0 & -12.21 \\
15 & Wave2D\_LongTime           & \textcolor{red}{\num{9.16e-1}} & 9.46e-1 & -3.26 \\
16 & KuramotoSivashinskyEquation& \textcolor{red}{\num{9.86e-1}} & 1.00e0 & -1.54 \\
17 & GrayScottEquation          & 2.61e2  & \textcolor{red}{\num{2.31e-1}} & \textbf{99.91} \\
18 & PoissonND                  & \textcolor{red}{\num{7.43e-3}} & 7.71e-3 & -3.81 \\
19 & HeatND                     & 1.04e-3 & \textcolor{red}{\num{9.14e-4}} & \textbf{12.05} \\
\hline
\end{tabular}
\end{table*}

Table \ref{tab:model_comparison}  quantitatively compared the performance of SEA-PINN and FNN-PINN on the 20 benchmark PDEs. It reports the averaged relative $L^2$ errors for 30 independent trials each starts with a different random seed and being trained for $5,000$ epochs.
Among the 20 cases, our SEA-PINN achieved lower errors in 13 cases, while FNN-PINN performed better in 7 cases. Specifically, SEA-PINN showed a significant improvement of over 5\% compared to FNN-PINN in 9 cases. However, SEA-PINN only exhibited a performance degradation of more than 5\% relative to FNN-PINN in two cases. In these two underperforming cases, the relative errors of both FNN-PINN and SEA-PINN were significantly larger than 100\% (that is, greater than 1 in relative error), indicating that neither network effectively learned the underlying patterns, rendering the comparison meaningless in these instances. Supplementary Fig. 14 visualizes the error distributions for FNN-PINN, SEA-PINN, and TSA-PINN; the violin plots of SEA-PINN are shorter in length than those of FNN-PINN in more than half of the cases, indicating lower error magnitude and greater stability in the final outcomes. 
SEA-PINN and TSA-PINN each outperform the other in 10 cases. In high-frequency problems where TSA-PINN excels, our method is not the top performer; nonetheless, compared to FNN-PINN it still achieves substantial gains—for example, in Heat2D\_Multiscale (case 7) and GrayScottEquation (case 17), our network improves by 83.09\% and 99.91\%, respectively. In non–high-frequency problems—particularly steady-state multi-output and steady-state multi-region scenarios—SEA-PINN exhibits consistent advantages. 
For the steady-state multi-output cases NS2D\_LidDriven (case 10) and NS2D\_BackStep (case 11), SEA-PINN achieves improvement percentages of 52.98\% and 80.90\%, respectively, notably surpassing TSA-PINN’s 41.60\% and 60.86\%. 
In the steady-state multi-region case Poisson2D\_ManyArea (case 5), SEA-PINN attains a 37.19\% improvement, compared with \(-1.01\%\) for TSA-PINN. 
Comparisons of more architectures are provided in Supplementary Table 3 and Supplementary Fig. 16.
In addition, by slightly increasing or decreasing the number of layers in the weight generator, we constructed two family variants. For each case, small-batch experiments can be used in the early stage to select the optimal weight generator, denoted SEA-PINN*. When comparing SEA-PINN* with the other two networks, it outperforms FNN-PINN in 15 out of 20 cases and TSA-PINN in 12 cases, demonstrating broader overall advantages (see Supplementary Table 4 and Supplementary Fig. 15).

Furthermore, we conducted tests and comparisons across a wider range of sampling point configurations, larger numbers of epochs, and more network architectures. After full training over 20,000 epochs, FNN-PINN, GAAF-PINN, LAAF-PINN, and our network were thoroughly evaluated. As shown in Supplementary Table 5, our method still demonstrates superior performance and achieves better accuracy.

\subsection{Superior Performance in Challenging Cases of Anisotropic Heat Conduction and Incompressible Fluid Dynamics}

\begin{figure*}[htbp]
\centering
\includegraphics[width=\linewidth,keepaspectratio]{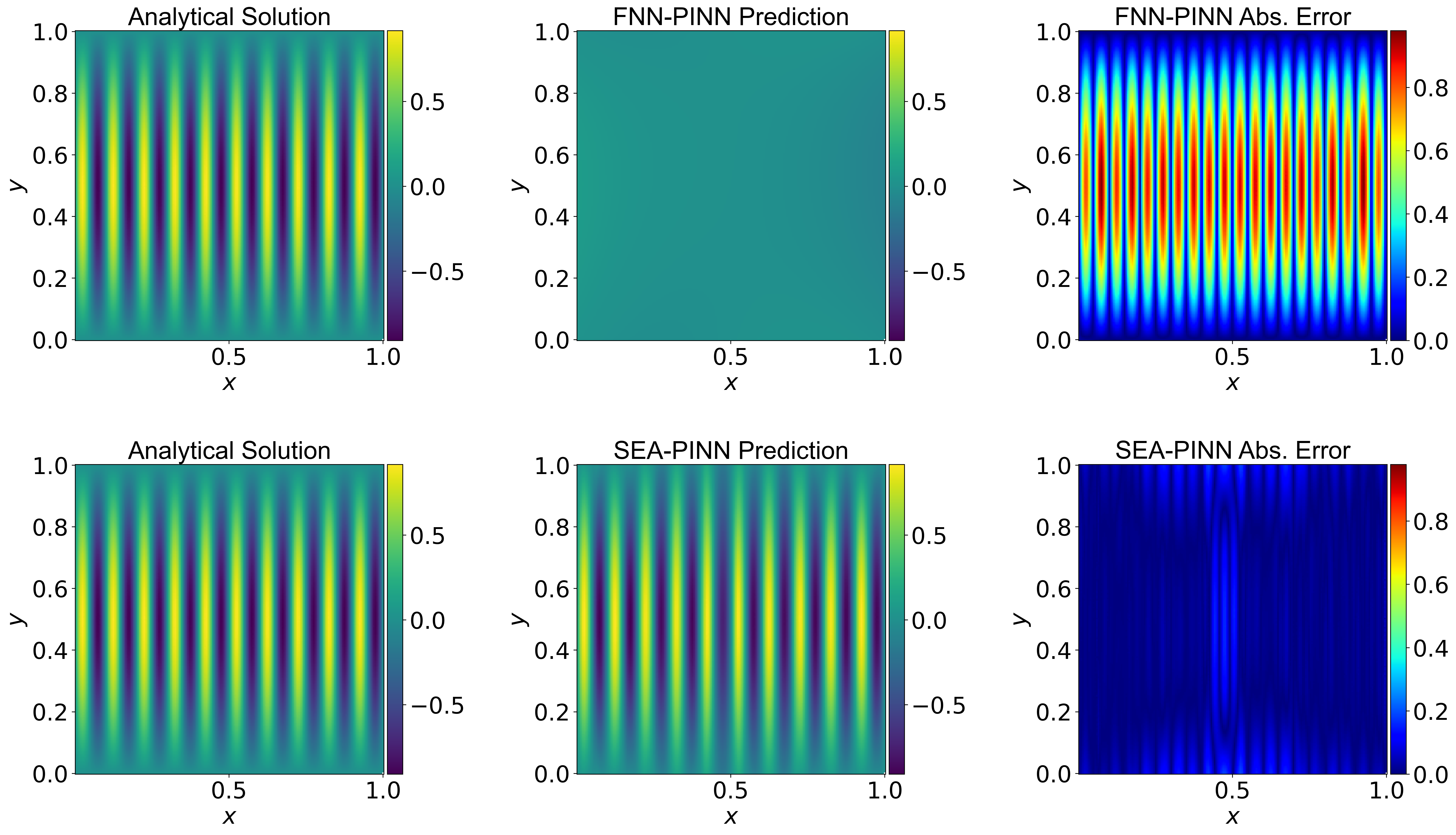}\\
\caption{
Heat2D\_ Multiscale(ID=7):Heatmaps of FNN-PINN and SEA-PINN at $t$=0.1s (Seed=1)}
\label{fig:heatmap_case7}
\end{figure*}

To rigorously evaluate the performance of SEA-PINN, we selected a challenging benchmark problem: a two-dimensional transient heat conduction problem with strong anisotropy and high-frequency spatial variations (Case 7). This case is representative of complex physical systems where traditional FNN-PINN often struggles. Additionally, Supplementary Figs. 12 and 13 show the heatmap comparisons for two steady multi-output Navier–Stokes benchmarks—a lid-driven cavity flow (Case 10) and a two-dimensional Navier–Stokes flow over a backward-facing step (Case 11). These results demonstrate that SEA-PINN maintains high accuracy across challenging problems with high spatial frequencies, and complex flow patterns.

The governing equation for case 7 is the 2D transient anisotropic heat equation:
\begin{align}
\frac{\partial u}{\partial t} = D_x \frac{\partial^2 u}{\partial x^2} + D_y \frac{\partial^2 u}{\partial y^2}
\label{eq:case7}
\end{align}
where $u(x, y, t)$ represents the temperature at spatial coordinates $(x, y) \in [0, 1]^2$ and time $t \in [0, 5]$. The diffusion coefficients are set to $D_x = \frac{1}{(500\pi)^2}$ and $D_y = \frac{1}{\pi^2}$. The system starts with an initial condition of $u(x, y, 0) = \sin(20\pi x)\sin(\pi y)$, and is subject to zero-Dirichlet boundary conditions on all spatial boundaries.

The diffusion coefficients creat a disparity of several orders of magnitude. The initial condition contains high-frequency components in the $x$-direction and low-frequency components in the $y$-direction. The challenge for the PINN model lies in capturing the vastly different dynamics simultaneously: the slow decay of high-frequency spatial features along the low-diffusivity $x$ axis and the faster evolution of smooth features along the high-diffusivity $y$ axis. This scenario tests the model's ability to handle high-frequency problems.

As illustrated in Fig.~\ref{fig:heatmap_case7}, we compare the temperature fields predicted by SEA-PINN and a standard FNN-PINN against the reference solution at early time. The results clearly demonstrate SEA-PINN's superiority. At early time ($t=0.1$s), SEA-PINN's prediction shows excellent agreement with the ground truth. In contrast, the FNN-PINN’s prediction is overly smooth and completely fails to resolve these fine-scale details, leading to a significant loss of accuracy. This highlights SEA-PINN's enhanced capability to model high-frequency physical phenomena.

\sect{Discussion}
The present study introduces the Squeeze-Excitation-like attention into Physics-Informed Neural Network (SEA-PINN), designed to address the persistent challenges of prediction accuracy and training stability in conventional Physics-Informed Neural Networks (PINNs). Our experimental results, based on a comprehensive benchmark comprising 20 diverse partial differential equations (PDEs), demonstrate that SEA-PINN significantly outperform traditional PINNs using Feedforward Neural Networks (FNNs) as backends. Moreover, in most cases, SEA-PINN also surpass the advanced TSA-PINN architecture, a robust baseline model that achieves higher accuracy in multiple challenging scenarios by incorporating neuron-wise sinusoidal activation functions with trainable frequencies and a dynamic slope recovery mechanism. Notably, SEA-PINN achieve substantially lower loss values with only a few training epochs  and exhibit significantly reduced variance across multiple runs on the same physical problem. Furthermore, without introducing Fourier transformations  or perodic activations, the SEA-PINN works well on high frequency solutions. These findings collectively suggest that SEA-PINN effectively mitigate key issues related to training stability and expressive power in existing PINN frameworks.

The superior performance of SEA-PINN implicate a fundamental differences in its representation power compared to standard FNNs. We have observed  much smaller mean and variance of the variational solution $u_{\theta}(t, \vec{x})$ and its derivatives of different orders in the PDEs, than traiditional FNN-PINN and TSA-PINN. 
Beyond these empirical observations, we further analyzed the output--input Jacobian and the layer-wise local Jacobians of the networks. 
Across 20 benchmark PDEs and multiple random seeds, SEA-PINN consistently exhibits smaller output--input gradient norms $\bigl\|\frac{\partial \mathbf{f}}{\partial \mathbf{x}}\bigr\|$ at initialization, as well as smaller Frobenius norms of the layer-wise Jacobians $\bigl\|\mathbf{D}_{\mathrm{SEA}}^{(l)}\bigr\|_F$ compared with FNN-PINN. 
In addition, the inter-layer condition numbers of the mapping from the first to the ninth hidden layer are substantially reduced in SEA-PINN compared with FNN-PINN.
Together, these results indicate that SEA-PINN realizes smoother and better-conditioned feature transformations, providing a more favorable starting point for optimization and helping to explain its lower initial loss, faster convergence, and reduced variance relative to FNN-PINN and TSA-PINN.

Experimental validation of this advantage is evident in our function approximation study (Supplementary Fig. 10), where SEA-PINN demonstrates superior extrapolation stability compared to standard networks. If the ground truth is $0$ in the external region, the SEA-PINN gives close-to-zero extrapolation with quite small variance. If the ground truth is not $0$, the SEA-PINN generate both smaller bias and smaller variance in 10 runs with different random seeds. This behavior directly translates to the observed lower initial PDE residuals and more consistent training dynamics across multiple random initializations, demonstrating the effectiveness of our approach for PDE-constrained optimization.

Initially, we explored symmetry-breaking mechanisms by introducing position embeddings to break the permutation symmetry of neurons in each layer. However, we found that dynamically generating scaling factors via a network not only yielded better performance but also implicitly constructed nonlinear connections between output neurons in each layer, making it more difficult for permutation symmetry to emerge. This approach aligns with the idea of attention, where the network learns to focus on relevant features or neurons adaptively.

The self-attention \cite{vaswani2017attention} widely used in Large Language Models (LLMs) is exceptionally powerful for tasks requiring long-range correlations, but comes at a significant cost, with computational complexity scaling quadratically with the sequence length. This has motivated several work on more efficient or structured attention mechanisms for long-context LLMs \cite{qin2024lightning, sun2023retentive}. Our squeeze-excitation-like attention uses a lightweight, trainable neural network known as the Weight Generator(WG) to map all neurons within a layer to a vector of neuron-specific weights. The additional computational cost is smaller than self-attention because of the squeeze operations in the SENet. 

Recent studies on gated attention in LLMs \cite{qiu2025gated, lin2025forgetting} have demonstrated that incorporating lightweight, head-specific multiplicative gates—particularly elementwise sigmoid gating applied to the outputs of Scaled Dot-Product Attention (SDPA)—can significantly enhance non-linearity, sparsity, and training stability. This gated attention mechanism can be expressed as $Y'=Y \odot \sigma(X W_{\theta})$, where $Y$ denotes the self-attention output derived from the previous layer's output $X$. In our SEA-PINN architecture, the corresponding formulation is $Y'=X\odot\sigma({\rm SqueezeExcitation}(X))$. This design explicitly enables SEA-PINN to capture intra-layer correlations and globally recalibrate the contributions of all neurons within a hidden layer. Since both approaches employ sigmoid activation for gating weights, SEA-PINN can be viewed as a form of gated attention. The gated attention in LLMs and SEA-PINN together provide complementary evidence that appropriately designed gating mechanisms can systematically improve stability and generalization across diverse neural architectures and scientific learning tasks.

Furthermore, the context-dependent modulation—where weights $\boldsymbol{\lambda}_{i}$ are conditioned on the global activation state $\mathbf{x}$—mimics how biological dendrites evaluate and modulate signals based on overall network activity \cite{larkum2004top, gidon2020dendritic}. This mechanism implements a form of gain modulation, often analogized to a "biological attention" mechanism, which enhances both stability and expressivity.

Despite these promising results, several limitations should be acknowledged. First, the benchmark problems, while diverse, may not fully represent all real-world scenarios. Second, the computational overhead of the scaling factor generator, though manageable, could be a concern for extremely large-scale problems. Future work should explore the scalability of SEA-PINN to higher-dimensional and more complex PDEs, as well as investigate the integration of other biologically plausible mechanisms to further enhance performance and robustness.

In summary, the SEA-PINN architecture represents a significant conceptual and practical advancement in Physics-Informed machine learning. By endowing neural networks with a dynamic, squeeze-excitation-like attention mechanism, it provides a robust and powerful solution to long-standing challenges in the field, paving the way for more accurate and reliable simulations of the physical world.

\begin{figure*}[!ht]
    \centering
    \makebox[\textwidth]{\includegraphics[width=\textwidth]{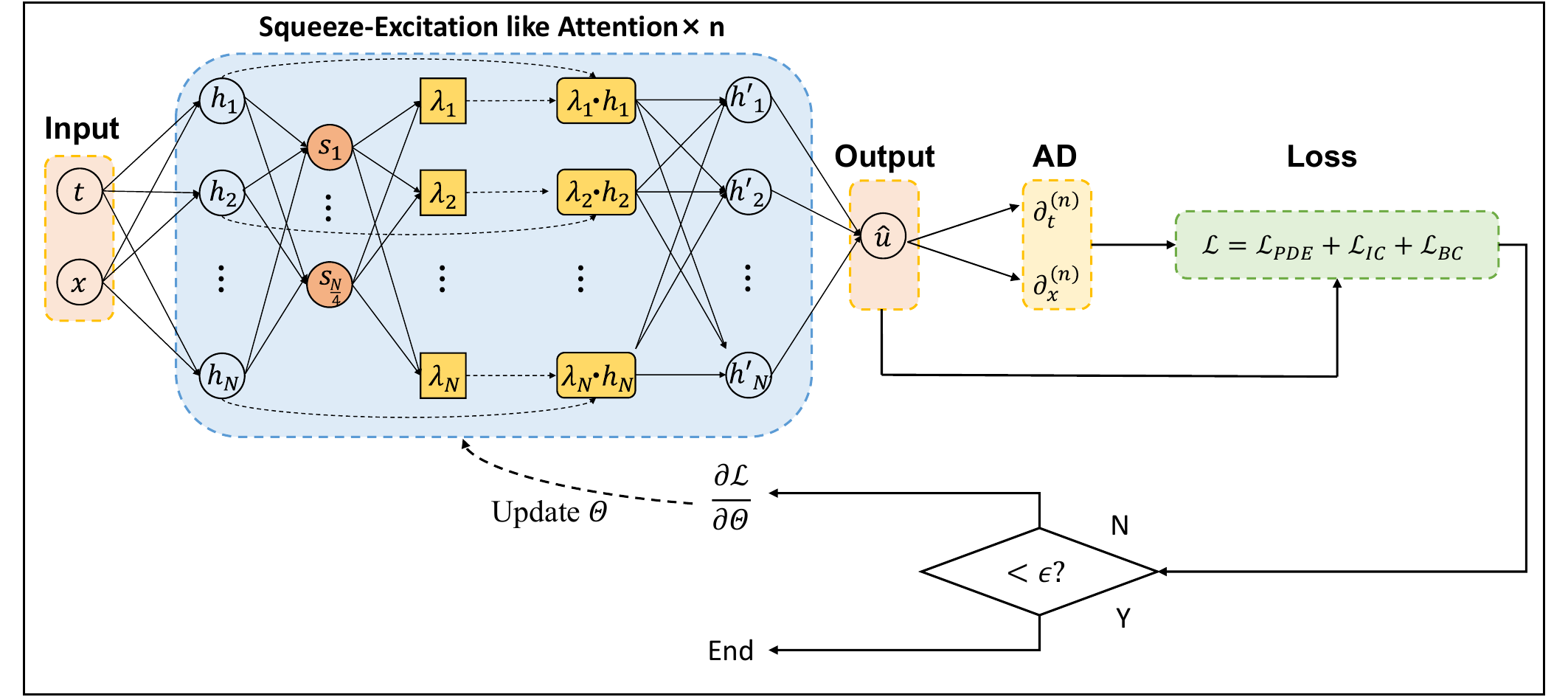}}
    \par\vspace{0.5\baselineskip} 
    \makebox[\textwidth]{
      \parbox{\textwidth}{
        \centering
        \caption{\raggedright 
        Architecture of SEA-PINN within PINNs: The complete set of outputs from the $n_{th}$ hidden layer is taken as input to the weight generator, producing a weight vector. These weights modulate the hidden activations through element-wise multiplication, yielding weighted activations, which are then fully connected to the $(n+1)_{th}$ hidden layer.}
        \label{fig:architecture}
      }
    }
  \end{figure*}
  
\sect{Method}

\subsection{Core Innovation: Physics-Informed Neural Network with Squeeze-Excitation-like Attention (SEA-PINN)}

We propose SEA-PINN, a Physics-Informed Neural Network enhanced with a Squeeze-Excitation-like Attention mechanism. As illustrated in Fig.~\ref{fig:architecture}, the model dynamically generates input-dependent weights for each neuron, thereby improving flexibility in solving PDEs. The forward computation at the $l$-th hidden layer proceeds through the following four steps:

1. Linear Transformation:
   \begin{align}
   \mathbf{z}^{(l)} = \mathbf{W}^{(l)}\mathbf{h}^{(l-1)} + \mathbf{b}^{(l)}
   \label{eq:LinearTransformation}
   \end{align}

2. Nonlinear Activation:
   \[
   \mathbf{a}^{(l)} = \sigma^{(l)}(\mathbf{z}^{(l)})
   \]
   where \( \mathbf{a}^{(l)} \in \mathbf{R}^{N_l} \).

3. Weight Generation (WG): The activation \( \mathbf{a}^{(l)} \) is processed by a lightweight squeeze-excitation block to produce a neuron-wise weight vector \( \boldsymbol{\lambda}^{(l)} \). This block consists of a small multilayer perceptron (MLP) with the sequence: \( \mathbf{a}^{(l)} \rightarrow \text{linear} + \text{Tanh} \rightarrow \text{linear} \rightarrow \text{Sigmoid} \rightarrow \boldsymbol{\lambda}^{(l)} \):
   
   \begin{align}
   \mathbf{z}_1 &= \mathbf{W}_1 \mathbf{a}^{(l)} + \mathbf{b}_1 \\
   \mathbf{z}_2 &= \mathbf{W}_2 \tanh(\mathbf{z}_1) + \mathbf{b}_2 \\
   \boldsymbol{\lambda}^{(l)} &= \sigma_{\text{Sigmoid}}(\mathbf{z}_2)
   \label{eq:WG}
   \end{align}
   
   Here, \( \mathbf{W}_1 \in \mathbf{R}^{\frac{N}{r} \times N} \) (with reduction ratio \( r=4 \)) compresses the input, while \( \mathbf{W}_2 \in \mathbf{R}^{N \times \frac{N}{r}} \) expands it back to the original dimension, thereby generating per-neuron weights.

4. Weighted Output:
   \begin{align}
   \mathbf{h}^{(l)} = \boldsymbol{\lambda}^{(l)} \odot \mathbf{a}^{(l)}
   \label{eq:WeightedOutput}
   \end{align}

This mechanism enables SEA-PINN to dynamically adjust the contribution of each neuron. If a neuron’s activation is less relevant for a specific input region in the current task, the weight generator can assign a smaller weight to suppress its influence; conversely, a larger weight can be assigned to enhance its role. This adaptive, neuron-level attention mechanism generates different $\mathbf{\lambda}^{(l)}$ for varying inputs, significantly enhancing the model’s flexibility and ability to approximate complex functions, such as PDE solutions.

\FloatBarrier

\subsection{Jacobian analysis of FNN-PINN and SEA-PINN}

Based on the layer-wise definitions in Eqs.~(\ref{eq:LinearTransformation})--(\ref{eq:WeightedOutput}),
we next analyze the output--input Jacobian of the baseline FNN and the proposed SEA-PINN,
as well as the corresponding layer-wise Jacobians. We denote the network input by
$\mathbf{x}$ and the network output by $\mathbf{f}$.

\paragraph{Baseline FNN.}
For a standard FNN without the weighting generator, the $l$-th hidden layer
($l=1,\dots,L-1$) reduces to the usual activation
\begin{equation}
    \mathbf{a}^{(l)}_{\mathrm{FNN}} = \sigma^{(l)}\!\bigl( \mathbf{z}^{(l)} \bigr),
\end{equation}
where $\mathbf{z}^{(l)}$ is given by Eq.~(\ref{eq:LinearTransformation}).
Since the nonlinearity is applied element-wise, the local Jacobian of
$\mathbf{a}^{(l)}_{\mathrm{FNN}}$ with respect to $\mathbf{z}^{(l)}$ is diagonal:
\begin{equation}
  \mathbf{D}_{\mathrm{FNN}}^{(l)}
  = \frac{\partial \mathbf{a}_{\mathrm{FNN}}^{(l)}}{\partial \mathbf{z}^{(l)}}
  = \operatorname{diag}\bigl( {\sigma^{(l)}}'(\mathbf{z}^{(l)})\bigr).
\end{equation}
The output layer is purely linear,
\begin{equation}
    \mathbf{f}_{\mathrm{FNN}} = \mathbf{W}^{(L)} \mathbf{a}^{(L-1)} + \mathbf{b}^{(L)} ,
\end{equation}
so the Jacobian of the network output with respect to the input $\mathbf{x}$ can be
written, using the chain rule, as
\begin{equation}
    \frac{\partial \mathbf{f}_{\mathrm{FNN}}}{\partial \mathbf{x}}
    =
    \mathbf{W}^{(L)}
    \prod_{l=1}^{L-1}
    \Bigl[
        \mathbf{D}_{\mathrm{FNN}}^{(l)} \mathbf{W}^{(l)}
    \Bigr].
    \label{eq:jacobian_fnn_seapinn}
\end{equation}

\paragraph{SEA-PINN.}
In SEA-PINN, each hidden layer ($l=1,\dots,L-1$) first computes the preliminary
activation $\mathbf{a}^{(l)} = \sigma^{(l)}(\mathbf{z}^{(l)})$ and then applies the
neuron-wise weights $\boldsymbol{\lambda}^{(l)} = \boldsymbol{\lambda}^{(l)}(\mathbf{a}^{(l)})$
generated by the WG module (Eq.~(\ref{eq:WG})). The final hidden representation used as
input to the next layer is given by the weighted output in Eq.~(\ref{eq:WeightedOutput}):
\begin{equation}
    \mathbf{h}_{\mathrm{SEA}}^{(l)} = \boldsymbol{\lambda}^{(l)} \odot \mathbf{a}^{(l)} 
\end{equation}
Using the product rule and writing $\boldsymbol{\lambda}^{(l)}$ as a function of
$\mathbf{z}^{(l)}$ for brevity, the local Jacobian with respect to $\mathbf{z}^{(l)}$ is
\begin{align}
    \mathbf{D}_{\mathrm{SEA}}^{(l)}
    &= 
    \frac{\partial \mathbf{h}_{\mathrm{SEA}}^{(l)}}{\partial \mathbf{z}^{(l)}} \notag\\
    &=
    \frac{\partial}{\partial \mathbf{z}^{(l)}}
    \bigl( \sigma^{(l)}(\mathbf{z}^{(l)}) \odot \boldsymbol{\lambda}^{(l)}(\mathbf{z}^{(l)}) \bigr) \notag\\
    &=
    \operatorname{diag}\bigl(\boldsymbol{\lambda}^{(l)}\bigr)
    \operatorname{diag}\bigl({\sigma^{(l)}}'(\mathbf{z}^{(l)})\bigr)
    +
    \operatorname{diag}\bigl(\sigma^{(l)}(\mathbf{z}^{(l)})\bigr)
    \frac{\partial \boldsymbol{\lambda}^{(l)}}{\partial \mathbf{z}^{(l)}} 
    \label{eq:D-sea}
\end{align}
Here $\operatorname{diag}(\cdot)$ maps a vector to a diagonal matrix and
$\frac{\partial \boldsymbol{\lambda}^{(l)}}{\partial \mathbf{z}^{(l)}} \in \mathbb{R}^{N_l\times N_l}$
denotes the Jacobian of the WG module. The output--input Jacobian of SEA-PINN is then
\begin{equation}
    \frac{\partial \mathbf{f}_{\mathrm{SEA}}}{\partial \mathbf{x}}
    =
    \mathbf{W}^{(L)}
    \prod_{l=1}^{L-1}
    \Bigl[
        \mathbf{D}_{\mathrm{SEA}}^{(l)} \mathbf{W}^{(l)}
    \Bigr]
    \label{eq:jacobian-sea}
\end{equation}

\paragraph{Layer-wise Jacobian norms.}
For each hidden layer $l$, we quantify the local sensitivity by the Frobenius norm of the
Jacobian,
\begin{equation}
  \begin{aligned}
    \bigl\|\mathbf{D}_{\mathrm{FNN}}^{(l)}\bigr\|_F
    &= \sqrt{\sum_{i=1}^{N_l} \bigl({\sigma^{(l)}}'(\mathbf{z}^{(l)}_i)\bigr)^2}, \\
    \bigl\|\mathbf{D}_{\mathrm{SEA}}^{(l)}\bigr\|_F
    &= \sqrt{\sum_{i,j=1}^{N_l}
        \bigl(\mathbf{D}_{\mathrm{SEA}}^{(l)}\bigr)_{ij}^2} .
  \end{aligned}
\end{equation}
In our experiments (Supplementary Fig. 11), we estimate these norms on real training points of the PINN for each layer, averaging over $1,000$ randomly selected samples and multiple random seeds. We observe that for both representative cases (Case~7 and Case~11)
and for all hidden layers, the SEA-PINN Jacobians satisfy
$\|\mathbf{D}_{\mathrm{SEA}}^{(l)}\|_F < \|\mathbf{D}_{\mathrm{FNN}}^{(l)}\|_F$ on average and also
exhibit smaller variance. Combined with the global Jacobian expressions
\eqref{eq:jacobian_fnn_seapinn}--\eqref{eq:jacobian-sea}, this indicates that SEA-PINN produces
smaller output--input gradient norms $\bigl\|\frac{\partial \mathbf{f}}{\partial \mathbf{x}}\bigr\|$,
leading to smoother initial PINN solutions and providing a better optimization starting point.

\subsection{Comparative Model: Trainable Sine Activation PINN}

We introduce the recently proposed Trainable Sine Activation PINN (TSA-PINN) as a key comparative model, representing an advanced architecture in the PINNs domain. This framework enhances the expressivity of traditional PINNs through neuron-level frequency modulation, demonstrating substantial improvements in fluid dynamics problems, such as those governed by the Navier-Stokes equations. Its core innovations consist of two components:

    The TSA mechanism replaces standard activation functions with a combination of sine and cosine functions with trainable frequencies:
\begin{align}
z_i^{(k)} &= \mathbf{w}_i^{(k)} \cdot \mathbf{a}^{(k-1)} + b_i^{(k)} \\
\psi_i^{(k)} &= \sin(f_i^{(k)} z_i^{(k)})\\
\phi_i^{(k)} &= \cos(f_i^{(k)} z_i^{(k)})\\
a_i^{(k)} &= \zeta_1 \psi_i^{(k)} + \zeta_2 \phi_i^{(k)} \notag \\
&= \zeta_1 \sin(f_i^{(k)} z_i^{(k)}) + \zeta_2 \cos(f_i^{(k)} z_i^{(k)})
\label{eq:TSAPINN1}
\end{align}
where:
\begin{itemize}
    \item $z_i^{(k)}$ is the output of the $i$-th neuron in the $k$-th layer,

    \item $f_i^{(k)} \in \mathbf{R}$ is an independent trainable frequency parameter for the $i$-th neuron in the $k$-th layer (default initialization: 1.0),

    \item $\mathbf{a}^{(k-1)} \in \mathbf{R}^{N_{k-1}}$ is the input vector from the ($k$-1)-th layer,

    \item $\mathbf{w}_i^{(k)} \in \mathbf{R}^{N_{k-1}}$ is the weight vector for the $i$-th neuron in the $k$-th layer,

    \item $b_i^{(k)} \in \mathbf{R}$ is the bias for the $i$-th neuron.
\end{itemize}
Meanwhile, TSA-PINN incorporates a slope recovery term into the loss function to rapidly enhance activation efficiency and accelerate the training process.

The complete mathematical formulation of both components is provided in Supplementary Information (Section: Trainable Sine Activation PINN) for reproducibility.

\subsection{Physics-Informed Neural Networks}

Physics-Informed Neural Networks (PINNs) are a deep learning framework that embeds physical laws, typically described by partial differential equations (PDEs), directly into the training process to solve differential equations. Unlike traditional data-driven neural networks, PINNs leverage physical constraints as a form of regularization, guiding the network to learn physically plausible solutions even in scenarios with sparse or no observational data.

The core concept of PINNs is to approximate the solution to a PDE using a deep neural network $u_{\theta}(t,\mathbf{x})$, where $\theta$ represents the trainable parameters weights and biases, and $t$ and $\mathbf{x}$ denote temporal and spatial coordinates, respectively. The PDE is expressed as:
\begin{align}
\mathcal{N}_u = f(x,t), \quad x \in \Omega, t \in [0,T]
\label{eq:PDE}
\end{align}
accompanied by initial conditions (ICs) and boundary conditions (BCs):
\begin{align}
u(x, 0) = g(x), \quad x \in \Omega
\label{eq:ICs}
\end{align}
\begin{align}
\mathcal{B}_u = h(x,t), \quad x \in \partial\Omega, t \in [0,T]
\label{eq:BCs}
\end{align}
where $u(x,t)$ is the target function, $\mathcal{N}[\cdot]$ is a linear or nonlinear differential operator, $\mathcal{B}[\cdot]$ is the boundary condition operator, $\Omega$ is the spatial domain, and $\partial\Omega$ is its boundary.

The training of PINNs is driven by minimizing a composite loss function $\mathcal{L}_{total}(\theta)$, which incorporates physical constraints, initial conditions, and boundary conditions. The loss function is defined as:
\begin{align}
\mathcal{L}_{total}(\theta) = w_{pde} \mathcal{L}_{pde}(\theta) + w_{ic} \mathcal{L}_{ic}(\theta) + w_{bc} \mathcal{L}_{bc}(\theta)
\label{eq:loss_function}
\end{align}
where $w_{pde}$, $w_{ic}$, and $w_{bc}$ are weights associated with the PDE residual, initial condition, and boundary condition losses, respectively. 

By minimizing $\mathcal{L}_{total}(\theta)$ using gradient-based optimization algorithms, PINNs learn a continuous and differentiable approximation of the PDE solution across the entire spatiotemporal domain.

\subsection{Experimental Setup}

\subsubsection{Benchmark Suite and Evaluation Metrics}

We evaluated the proposed models on 20 challenging PDE problems selected from a recognized PINNs benchmark suite, covering diverse domains including heat conduction, fluid dynamics, biology, electromagnetics, and high-dimensional problems \cite{zhongkai2024pinnacle}. This ensures a comprehensive assessment of model performance. As all cases are forward problems, model training relies solely on physical constraints (i.e., PDE equations, initial conditions, and boundary conditions) without involving any measured data from true solutions. This approach ensures a fair evaluation of the models’ ability to solve PDEs in purely physics-driven scenarios.

To account for randomness, each problem is trained and tested with 30 random seeds, enabling reproducible results and averaging to reduce variability. We report the relative $L^2$ error defined on the evaluation set as
$E_{\text{rel}} = \frac{\lVert \mathbf{y} - \mathbf{y}_{\text{ref}} \rVert_2}{\lVert \mathbf{y}_{\text{ref}} \rVert_2}$,
where $\lVert \cdot \rVert_2$ denotes the Euclidean norm over all sampled values; for vector-valued fields, components are concatenated before computing the norm.

\subsubsection{Network Configuration and Training}
	
	For a fair comparison, all neural network models (FNN-PINN, SEA-PINN, TSA-PINN, TSA\_SEA-PINN) share the same backbone architecture, consisting of 9 hidden layers with 32 neurons per layer. Unless otherwise specified, all weights are initialized using He uniform initialization and biases are set to zero.
	
	\textbf{Activation Functions:} 
	
	\begin{itemize}
		
		\item Backbone network for FNN-PINN and SEA-PINN: SiLU.
		
		\item Weight generator in SEA-PINN: Tanh.
		
	\end{itemize}
	
	\textbf{Optimizer and Learning Rate:} Adam optimizer with an initial learning rate of $10^{-3}$ \cite{kingma2014adam}. 
	
	\textbf{Training Process:} Each model is trained for 5{,}000 epochs using full-batch training (no mini-batching).

    \textbf{Training Data Sampling:} The number of collocation points used for training is kept consistent across all experiments. Specifically, for each training process, we sample 8,192 points for the PDE residual loss in the domain (\textit{domain points}) and 2,048 points for the boundary condition loss (\textit{boundary points}). For time-dependent problems, an additional 2,048 points are sampled for the initial condition loss (\textit{initial points}). The sampling of all points is conducted using the Hammersley sequence, a quasi-random low-discrepancy sequence that ensures a more uniform coverage of the solution domain compared to standard pseudo-random sampling. A fixed sampling strategy is employed, meaning the set of collocation points is generated once before training and is not resampled during the epochs.
	
	  \textbf{Loss Weights:} All loss terms are equally weighted ($w_{pde} = w_{ic} = w_{bc} = 1.0$).
      
    \textbf{Datasets:} For cases without analytical solutions (cases with analytical solutions are: case 7, 13, 15, 18, 19), reference solutions are generated using high-precision numerical methods. High-precision numerical solutions are obtained via the FEM solver through COMSOL 6.0  provided by \cite{zhongkai2024pinnacle} for problems with complex geometries and the spectral method provided by Chebfun \cite{driscoll2014chebfun} for chaotic problems.

	These settings are kept consistent across all models to ensure that any performance difference arises from the architectural innovations rather than training hyperparameters.

Supplementary Information: detailed training loss analysis, total training loss comparison, individual loss component comparison, initial loss behavior of SEA-PINN components, foundational performance on function approximation, layer-wise Jacobian norms for representative PDE cases, heatmap, comparison of relative $L^2$ errors across different architectures, the comparison of SEA-PINN variants, extended experimental evaluation, trainable sine activation PINN, hybrid architecture: TSA\_SEA-PINN, enhancing TSA-PINN with SEA-PINN: demonstrating superior synergistic performance, and total training loss are provided in the SI.
\FloatBarrier 

\section*{Declarations}
\begin{description}
    \item[Competing interests] All authors declare no competing interests.
    \item[Code availability] The source code and representative reproducibility examples are available at \url{https://github.com/YunFei-Song/SEA-PINN}.
    \item[Author contribution]\hspace{0pt} \\
    Yun-Fei Song was responsible for the methodology, software, investigation, visualization and writing the original draft. Long-Gang Pang was responsible for conceptualization, methodology, software, formal analysis, reviewing and editing the manuscript, supervision, and funding acquisition. Fu-Peng Li was responsible for reviewing and editing the manuscript. Jun-Jie Zhang was responsible for conceptualization, formal analysis, reviewing and editing the manuscript.

\end{description}

\begin{acknowledgments}
    This work is supported by the National Natural Science Foundation of China under Grant No.\ 12075098, No.\ 12435009. The work is also partly supported by the Fundamental Research Funds for the Central Universities at CCNU under Grant No.\ CCNU24JC011, the Cross Research Project of Fundamental Research Funds for Central Universities of Central China Normal University in 2025: "Advanced Detection and
 Artificial Intelligence at the Frontiers of Physics"(No.30101250317). The numerical calculation have been performed on the GPU cluster in the Nuclear Science Computing Center (NSC3) at CCNU. 
 \end{acknowledgments}

\bibliographystyle{apsrev4-1}  
\bibliography{ref}

\clearpage
\onecolumngrid
\rowcolors{2}{}{lightbrown!80}

\setcounter{section}{0}
\setcounter{subsection}{0}
\setcounter{figure}{0}
\setcounter{table}{0}
\setcounter{equation}{0}

\renewcommand{\thesection}{S\arabic{section}}
\renewcommand{\thesubsection}{S\arabic{section}.\arabic{subsection}}
\renewcommand{\thefigure}{S\arabic{figure}}
\renewcommand{\thetable}{S\arabic{table}}
\renewcommand{\theequation}{S\arabic{equation}}

\renewcommand{\theHsection}{S.\arabic{section}}
\renewcommand{\theHsubsection}{S.\arabic{section}.\arabic{subsection}}
\renewcommand{\theHfigure}{S.\arabic{figure}}
\renewcommand{\theHtable}{S.\arabic{table}}
\renewcommand{\theHequation}{S.\arabic{equation}}

\makeatletter
\renewcommand{\fnum@table}{\textbf{Supplementary Table~\thetable}}
\renewcommand{\fnum@figure}{\textbf{Supplementary Figure~\thefigure}}
\makeatother

\newcolumntype{L}{>{\RaggedRight}X}

\begin{center}
{\Large \bfseries Supplementary Information\par}
\vspace{0.8em}
\end{center}

\vspace{1em}

\section{Detailed Training Loss Analysis}
	
	This document provides supplementary results to support the claims made in the main text, demonstrating the consistent and significant advantages of the SEA-PINN over the standard FNN-PINN. We present detailed training loss comparisons for five representative benchmark cases, selected to cover a diverse range of PDE types and complexities.

	\begin{table}[htbp] 
		\centering 
		\caption{Summary of the five representative benchmark cases.} 
		\label{tab:cases_summary} 
		\small 
		\begin{tabular}{@{} l l l l @{}} 
			\toprule[1.1pt] 
			\textbf{Case} & \textbf{Governing Equation} & \textbf{Domain} & \textbf{Boundary / Initial Condition} \\ 
			\midrule[0.75pt] 
			Case 1 & 
			\parbox[t]{5.5cm}{\raggedright  2D Coupled Burgers equation: \\$\mathbf{u}_t + \mathbf{u}\!\cdot\!\nabla \mathbf{u} - \nu   \Delta \mathbf{u} = \mathbf{0}$ \\
            } & 
			\parbox[t]{3.5cm}{\raggedright $(x,y,t)\in \Omega = [0,L]^2\times[0,T]$ with $L=4$, $T=1$} & 
			\parbox[t]{5.5cm}{\raggedright
            $\boldsymbol{u}(0, y, t) = \boldsymbol{u}(L, y, t)$, \\
            $\boldsymbol{u}(x, 0, t) = \boldsymbol{u}(x, L, t)$ \\ ${x,y} \in[0,L], t \in [0,T]$}\\
			\addlinespace[3pt] 
			Case 5 & 
			\parbox[t]{5.5cm}{\raggedright 2D Poisson equation with many subdomains:\\
            $-\nabla (a(x) \nabla \mathbf{u}) = f(x, y)$} & 
			\parbox[t]{3.5cm}{\raggedright $(x,y) \in \Omega=[-10,10]^2$} & 
			\parbox[t]{5.5cm}{\raggedright$\partial\Omega$: $\displaystyle \frac{\partial \mathbf{u}}{\partial n} + \mathbf{u} = 0, x \in \partial \Omega$.} \\ 
			\addlinespace[3pt] 
			Case 7 & 
			\parbox[t]{5.5cm}{\raggedright  2D Heat Multi-Scale: $\displaystyle \frac{\partial u}{\partial t} - \frac{1}{(500\pi)^2} u_{xx}-\frac{1}{\pi^2}u_{yy}=0$} & 
			\parbox[t]{3.5cm}{\raggedright $\Omega \times T = [0,1]^2\times[0,5]$} & 
			\parbox[t]{5.5cm}{\raggedright $u(x,y,0)=\sin(20\pi x)\sin(\pi y), x \in \Omega$,\\
            $u(x,y,t) = 0, x \in \partial \Omega$.} \\ 
			\addlinespace[3pt] 
			Case 11 & 
			\parbox[t]{5.5cm}{\raggedright 2D Back Step Flow: \\$\mathbf{u}\!\cdot\!\nabla\mathbf{u} + \nabla p - \frac{1}{Re} \Delta \mathbf{u} = 0$,$Re=100$\\ $\nabla\!\cdot \mathbf{u}=0$} & 
			\parbox[t]{3.5cm}{\raggedright  $\Omega = [0,4]\times[0,2] \setminus \left( [0,2] \times [1,2] \cup R_i \right)$ (excluding the top-left quarter)} & 
			\parbox[t]{5.5cm}{\raggedright Inlet ($x{=}0$): $u(y)=4y(1-y)$, $v=0$; outlet ($x{=}4$): $p=0$; no-slip walls elsewhere: $u=v=0$.} \\ 
			\addlinespace[3pt] 
			Case 14 & 
			\parbox[t]{5.5cm}{\raggedright 2D Wave Equation in Heterogeneous Medium: \\$[\nabla^2 - \frac{1}{c(x)}\frac{\partial^2 }{\partial t^2}] u(x,t) =0$ (spatially varying $c$ from coefficient field)} & 
			\parbox[t]{3.5cm}{\raggedright $\Omega \in[-1,1]^2$} & 
			\parbox[t]{5.5cm}{\raggedright  $u(x,y,0)=\exp\!\big(-\tfrac{(x-\mu_x)^2+(y-\mu_y)^2}{2\sigma^2}\big),(x,y) \in \Omega$,\\ $\frac{\partial u}{\partial t} (x,y,0)=0, (x,y) \in \Omega$,\\ $\frac{\partial u}{\partial n} =0, (x,y) \in \Omega$.} \\ 
			\bottomrule[1.1pt] 
		\end{tabular} 
	\end{table}

	Table \ref{tab:cases_summary} provides a self-contained summary of the governing equations, domains, and boundary/initial conditions for these five cases.

	\sect{Total Training Loss Comparison}
	
\begin{figure}[!htbp] 
			\centering
			\begin{tabular}{c}
				\includegraphics[width=1\linewidth,keepaspectratio]{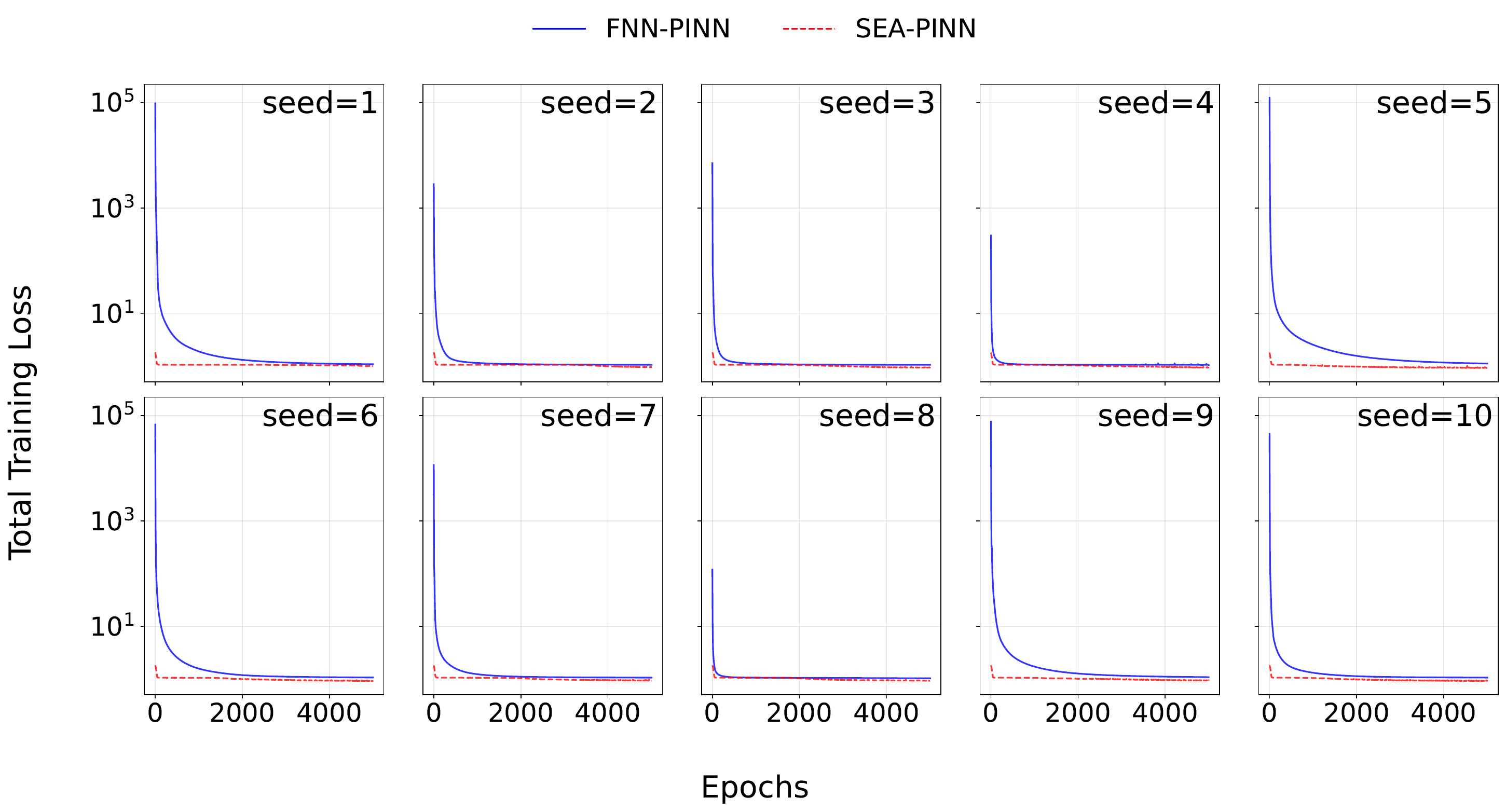}
			\end{tabular}
			\caption{Total training loss comparison for Burgers2D (Case 1) across 10 seeds}
			\label{fig:Total training loss comparison for Burgers2D(Case 1) across 10 seeds}
		\end{figure}

\begin{figure}[!htbp] 
			\centering
			\begin{tabular}{c}
				\includegraphics[width=1\linewidth,keepaspectratio]{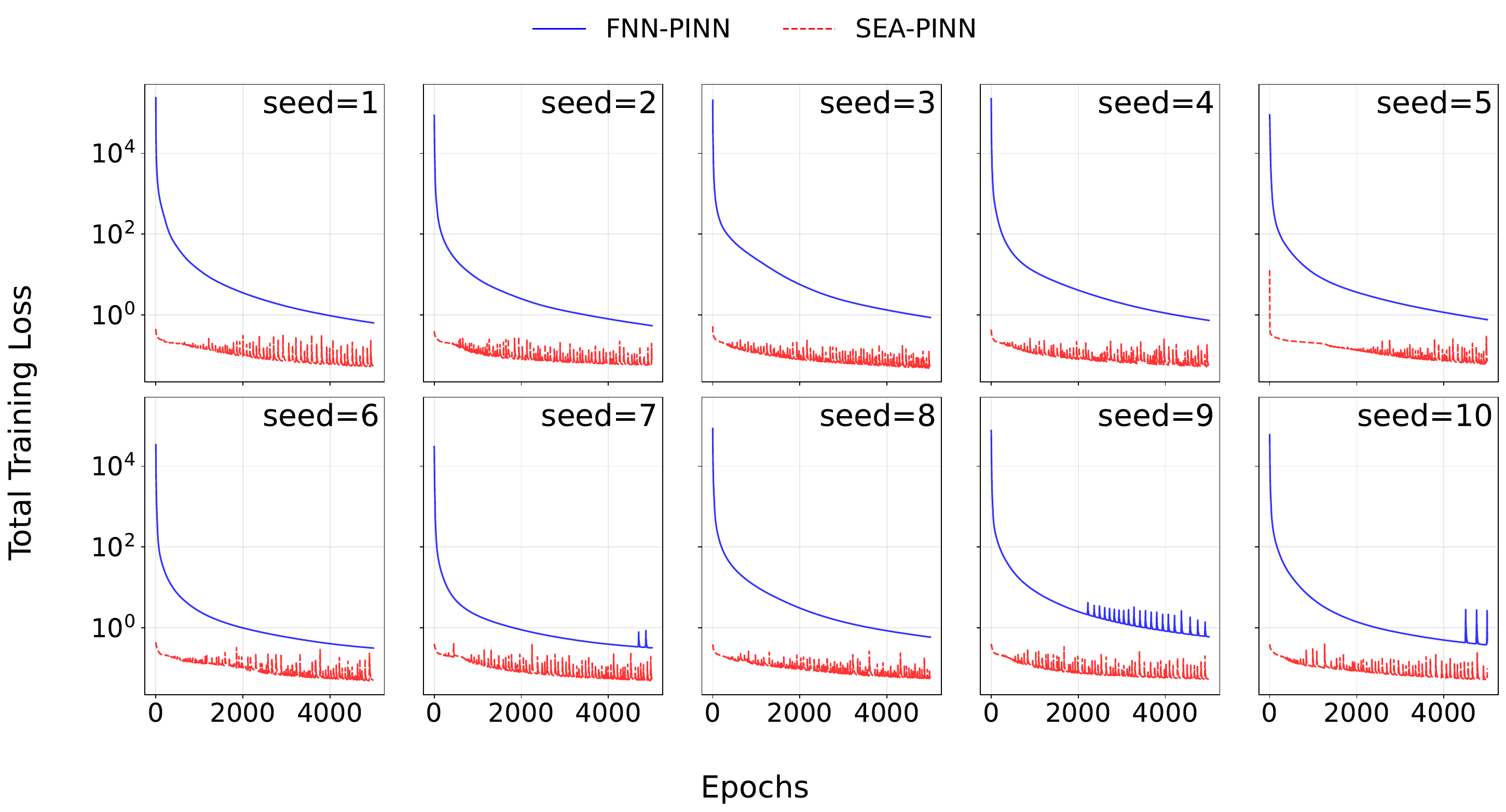}
			\end{tabular}
			\caption{Total training loss comparison for Poisson2D\_ManyArea (Case 5) across 10 seeds}
			\label{fig:Total training loss comparison for Poisson2D_ManyArea(Case 5) across 10 seeds}
		\end{figure}

\begin{figure}[!htbp] 
			\centering
			\begin{tabular}{c}
				\includegraphics[width=1\linewidth,keepaspectratio]{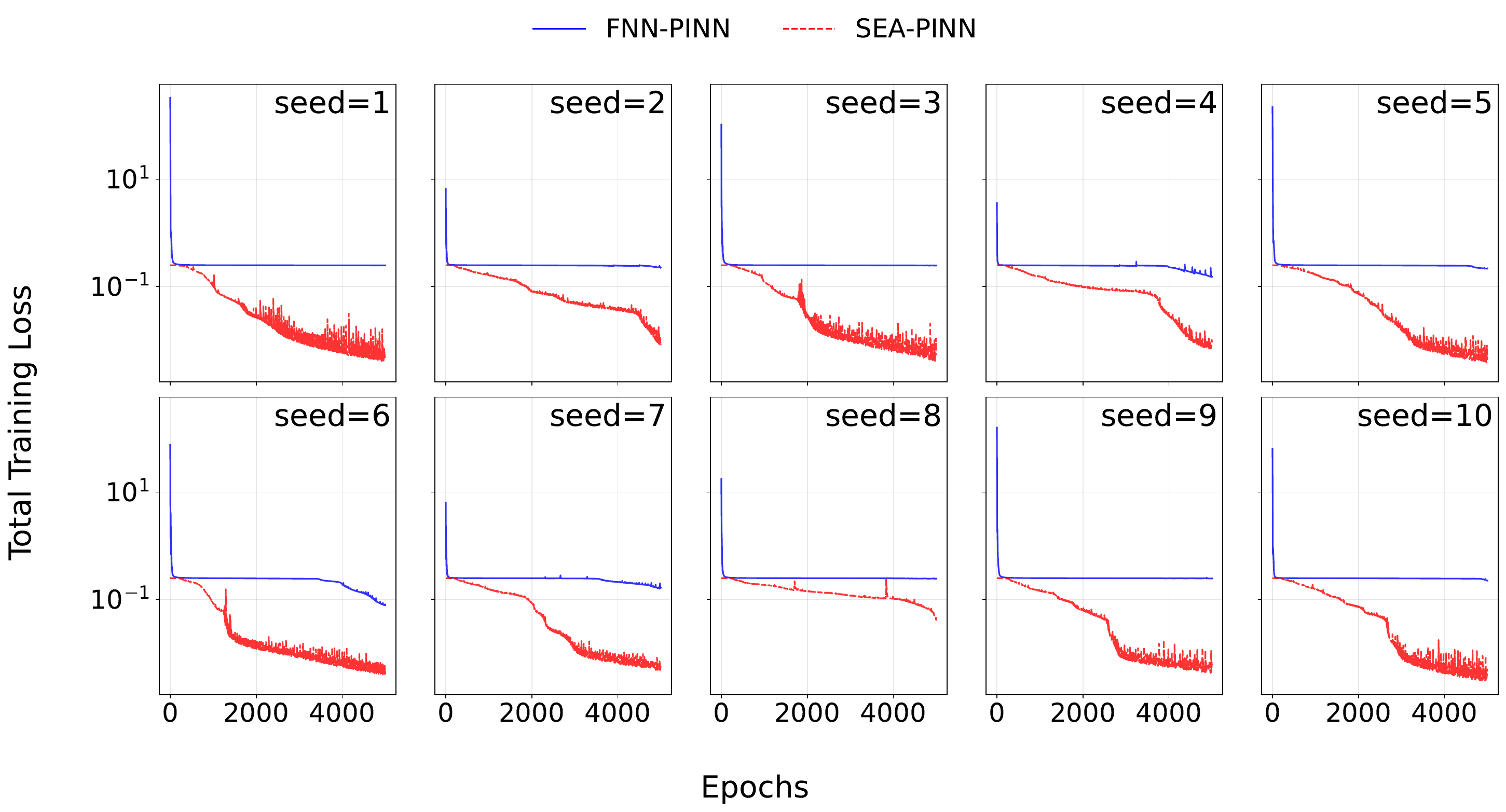}
			\end{tabular}
			\caption{Total training loss comparison for Heat2D\_Multiscale (Case 7) across 10 seeds}
			\label{fig:Total training loss comparison for Heat2D_Multiscale(Case 7) across 10 seeds}
		\end{figure}

\begin{figure}[!htbp] 
			\centering
			\begin{tabular}{c}
				\includegraphics[width=1\linewidth,keepaspectratio]{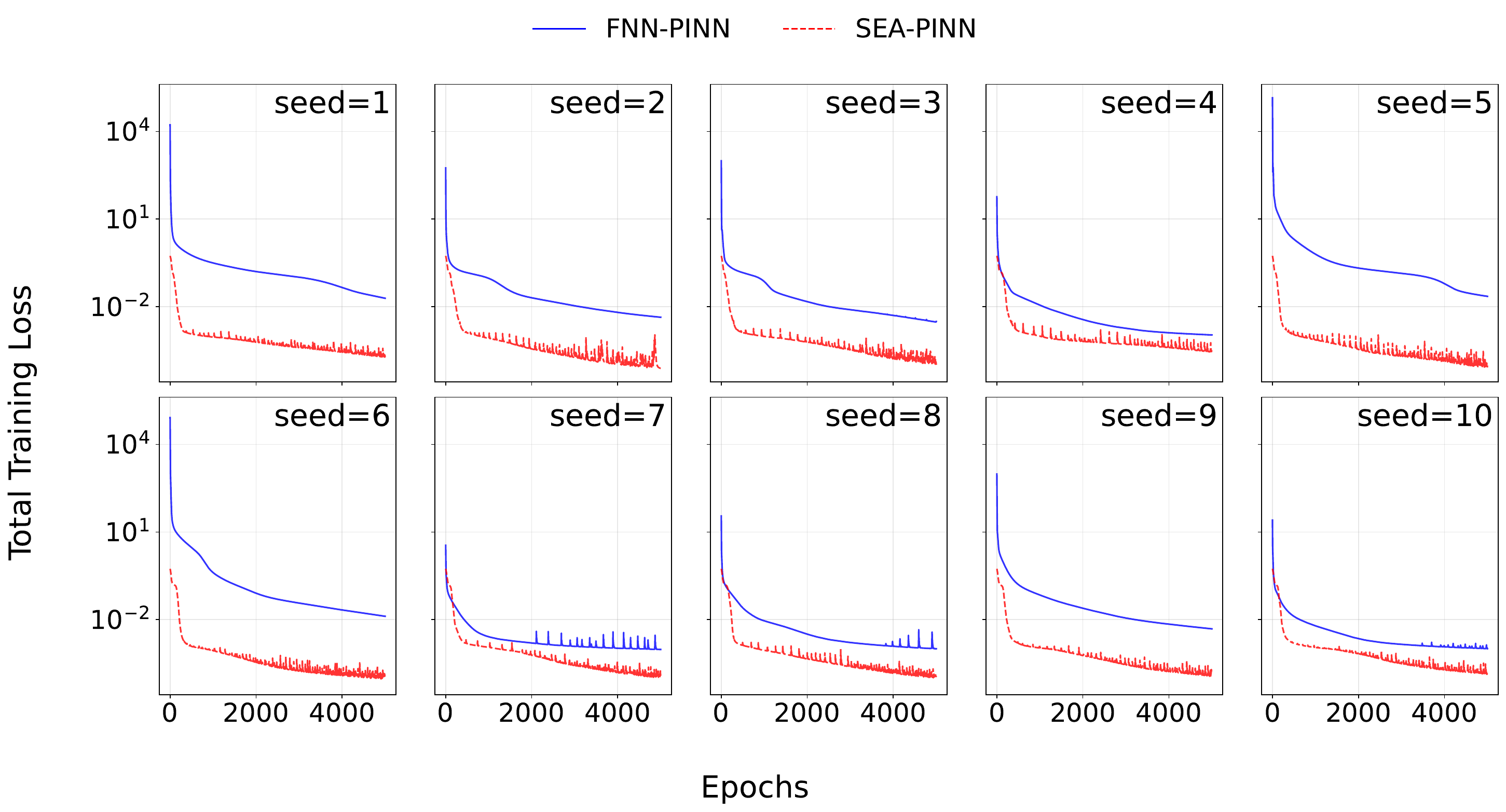}
			\end{tabular}
			\caption{Total training loss comparison for NS2D\_BackStep (Case 11) across 10 seeds}
			\label{fig:Total training loss comparison for NS2D_BackStep(Case 11) across 10 seeds}
		\end{figure}

\begin{figure}[!htbp] 
			\centering
			\begin{tabular}{c}
				\includegraphics[width=1\linewidth,keepaspectratio]{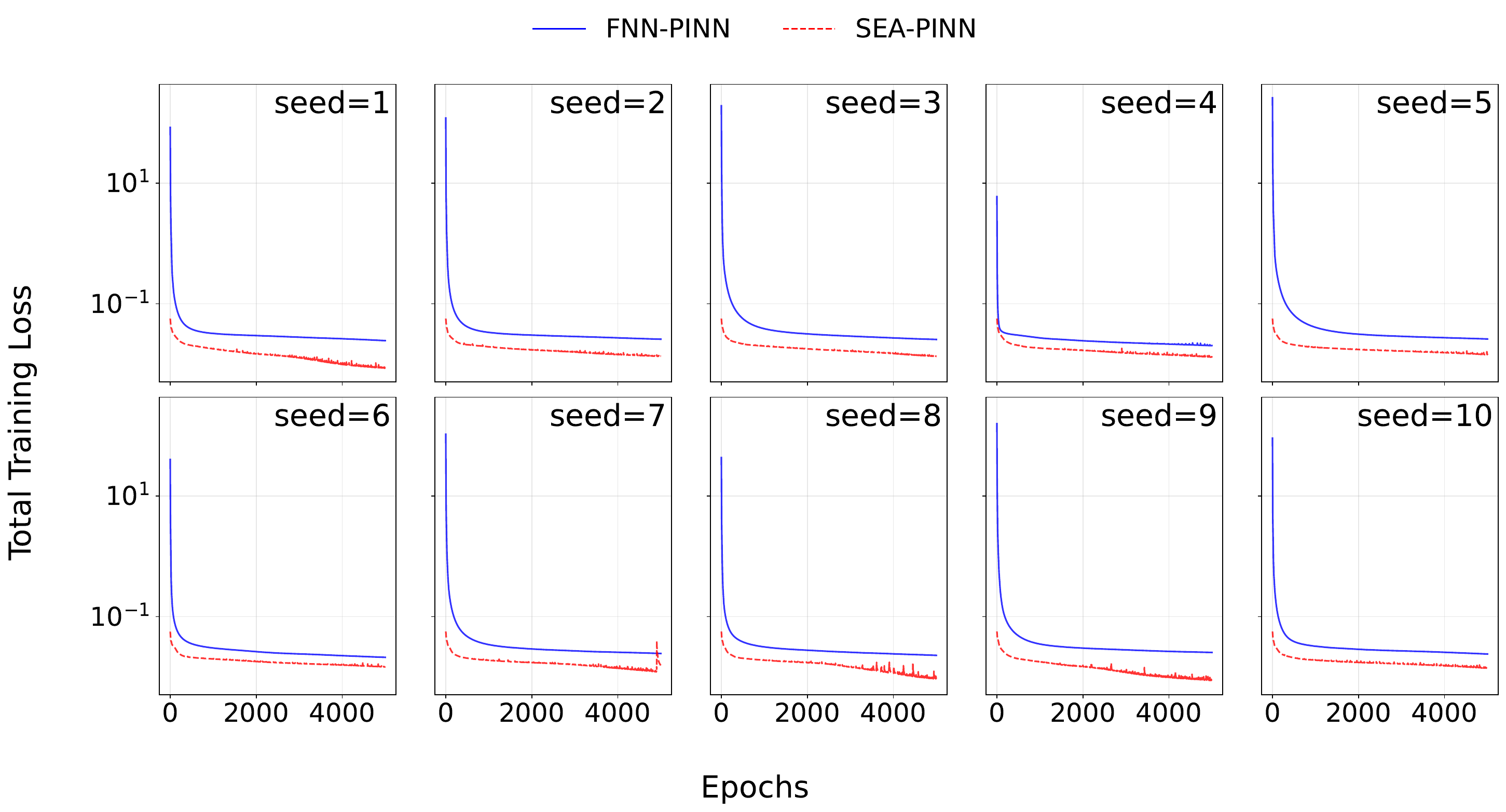}
			\end{tabular}
			\caption{Total training loss comparison for Wave2D\_Heterogeneous (Case 14) across 10 seeds}
			\label{fig:Total training loss comparison for Wave2D_Heterogeneous(Case 14) across 10 seeds}
		\end{figure}

	To demonstrate the robustness and consistent performance of SEA-PINN, we trained both SEA-PINN and FNN-PINN on each benchmark case using 10 different random seeds(seed:1-10). Figs. \ref{fig:Total training loss comparison for Burgers2D(Case 1) across 10 seeds}-\ref{fig:Total training loss comparison for Wave2D_Heterogeneous(Case 14) across 10 seeds} show the evolution of the total training loss over 5,000 epochs. Across all cases, SEA-PINN consistently converges to a significantly lower loss value and exhibits a more stable training trajectory compared to FNN-PINN, underscoring its superior optimization efficiency.

    \rowcolors{1}{}{}
    \normalcolor
		
        	\FloatBarrier

	\sect{Individual Loss Component Comparison}
	
\begin{figure}[!htbp] 
			\centering
			\begin{tabular}{c}
				\includegraphics[width=1\linewidth,keepaspectratio]{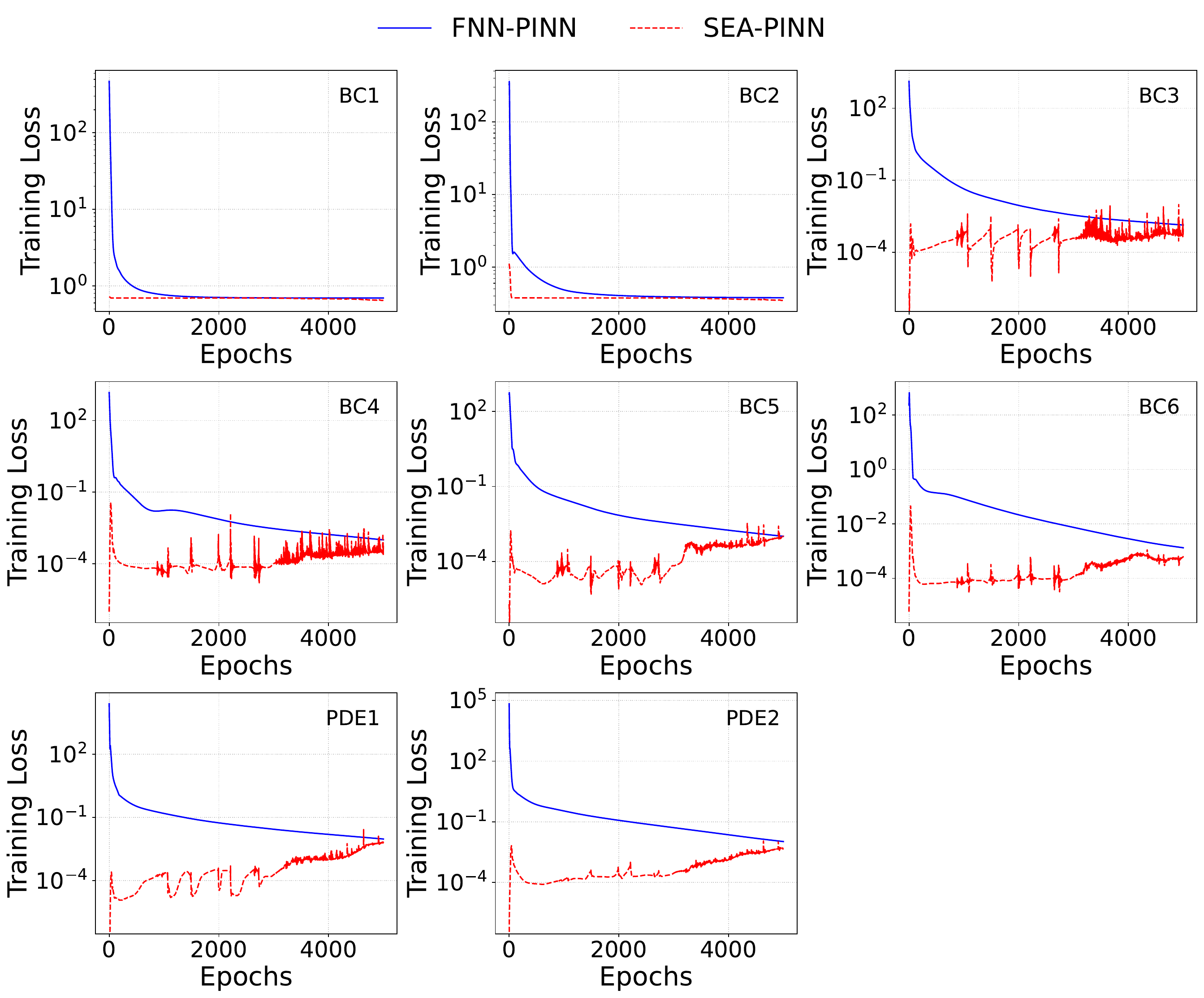}
			\end{tabular}
			\caption{Comparison of loss components for Burgers2D (Case 1)}
			\label{fig:Comparison of loss components for Burgers2D (Case 1)}
		\end{figure}

\begin{figure}[!htbp] 
			\centering
			\begin{tabular}{c}
				\includegraphics[width=1\linewidth,keepaspectratio]{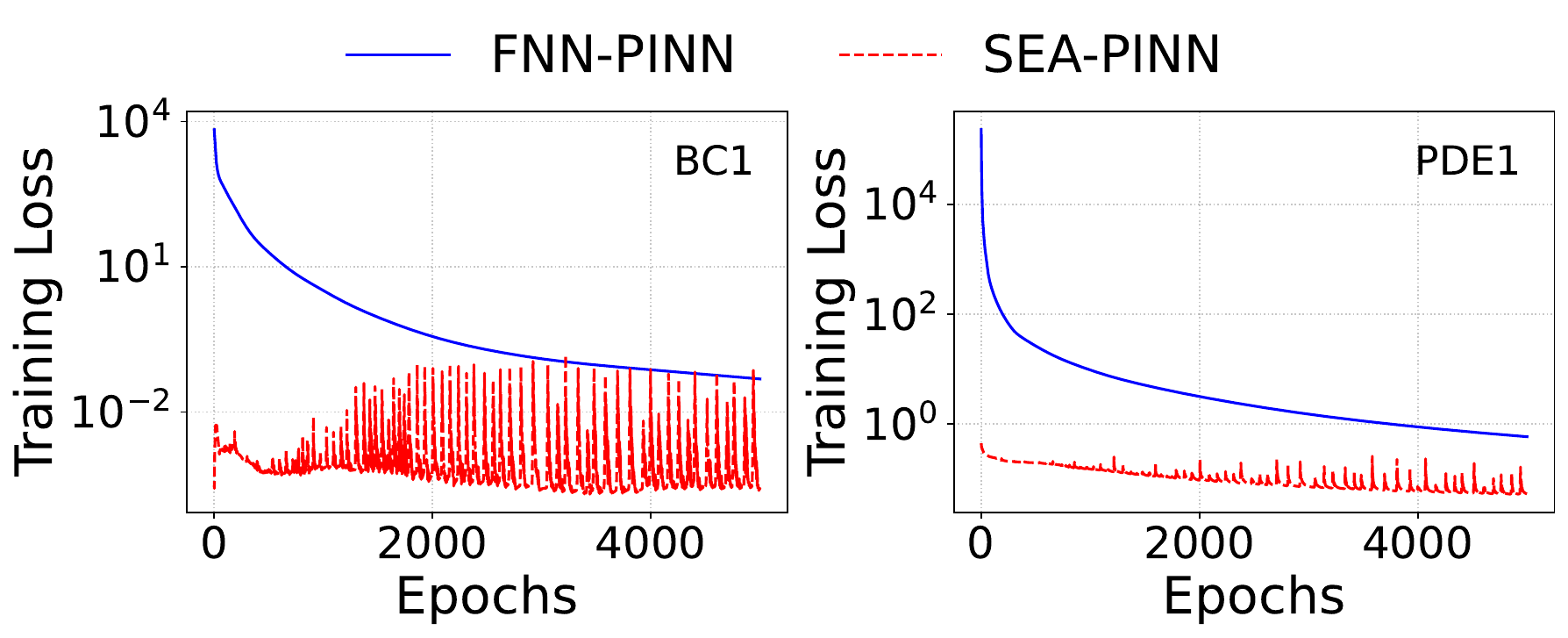}
			\end{tabular}
			\caption{Comparison of loss components for Poisson2D\_ManyArea (Case 5)}
			\label{fig:Comparison of loss components for Poisson2D_ManyArea (Case 5)}
		\end{figure}

\begin{figure}[!htbp] 
			\centering
			\begin{tabular}{c}
				\includegraphics[width=1\linewidth,keepaspectratio]{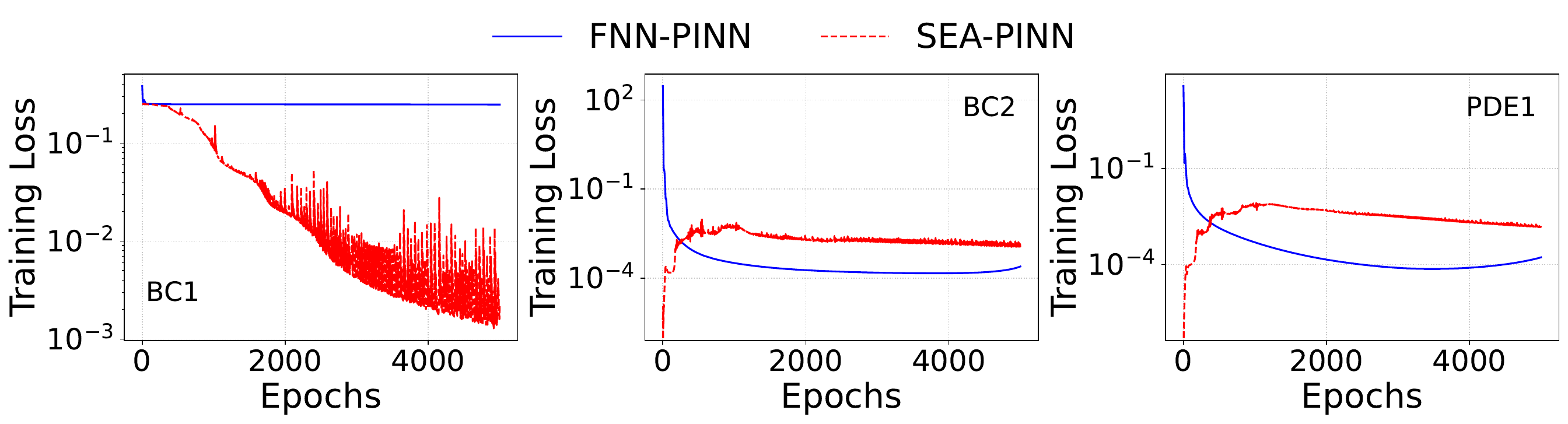}
			\end{tabular}
			\caption{Comparison of loss components for Heat2D\_Multiscale (Case 7)}
			\label{fig:Comparison of loss components for Heat2D_Multiscale(Case 7)}
		\end{figure}

\begin{figure}[!htbp] 
			\centering
			\begin{tabular}{c}
				\includegraphics[width=1\linewidth,keepaspectratio]{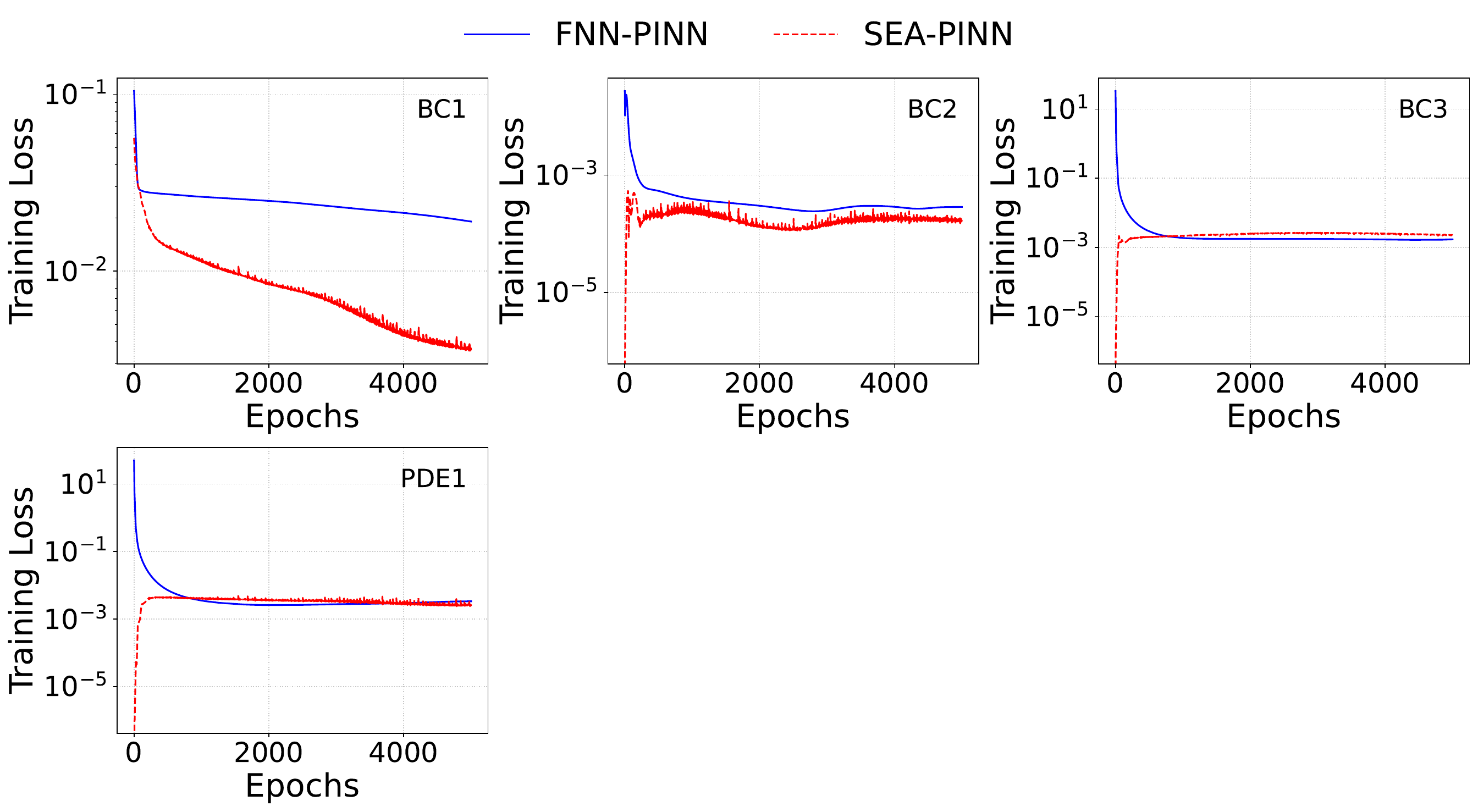}
			\end{tabular}
			\caption{Comparison of loss components for Wave2D\_Heterogeneous (Case 14)}
			\label{fig:Comparison of loss components for Wave2D_Heterogeneous(Case 14)}
		\end{figure}

	To provide a more granular view of the optimization dynamics, we compare the individual loss components (PDE residuals and boundary/initial condition losses) for a single training run (seed=1). For conciseness in the figure legends, losses from both boundary and initial conditions are collectively labeled 'BC', reflecting their unified treatment as constraints within our framework. As shown in Figs. \ref{fig:Comparison of loss components for Burgers2D (Case 1)}-\ref{fig:Comparison of loss components for Wave2D_Heterogeneous(Case 14)}, SEA-PINN's advantages are often established at the very beginning of the training. It is noteworthy that for many components, SEA-PINN's initial loss is several orders of magnitude lower than FNN-PINN's. This suggests that SEA-PINN provides a much better starting point for the optimization, allowing the model to find a solution that satisfies the physical constraints more easily.

     \FloatBarrier

\sect{Initial Loss Behavior of SEA-PINN Components}

\rowcolors{2}{}{lightbrown!80} 

\begin{table}[htbp]
\centering
\caption{Initial Loss Behavior of SEA-PINN Components for 20 Test Cases}
\label{tab:loss_behavior}
\renewcommand{\arraystretch}{1.3}
\scriptsize
\setlength{\tabcolsep}{2.5pt}
\begin{adjustbox}{max width=\textwidth} 
\begin{tabular}{ccccccc}
\toprule
\rowcolor{white}
\textbf{Case ID} & 
\textbf{Zero BC/IC} & 
\textbf{Low Initial Loss} & 
\textbf{Non-zero BC/IC} & 
\textbf{High Initial Loss} & 
\textbf{Ambiguous Trend} & 
\textbf{Remarks} \\
\midrule
0 & bc\_2 & bc\_2,pde\_1 & ic\_1 & ic\_1 &  &  \\
1 &  & \begin{tabular}[c]{@{}c@{}}bc\_3,bc\_4,bc\_5,bc\_6,\\pde\_1,pde\_2\end{tabular} & \begin{tabular}[c]{@{}c@{}}ic\_1,ic\_2\end{tabular} & \begin{tabular}[c]{@{}c@{}}ic\_1,ic\_2\end{tabular} &  & \begin{tabular}[c]{@{}c@{}}bc\_3,bc\_4,bc\_5,bc\_6\\is periodic\end{tabular} \\
2 & bc\_2 & bc\_2,pde\_1 & bc\_1 & bc\_1 &  &  \\
3 &  & bc\_1 & \begin{tabular}[c]{@{}c@{}}bc\_1,bc\_2\end{tabular} & pde\_1 & bc\_2 & bc\_1=0.2 is small \\
4 & bc\_1 & bc\_1 &  & pde\_1 &  &  \\
5 &  &  & bc\_1 & pde\_1 & bc\_1 & \begin{tabular}[c]{@{}c@{}}bc\_1 is robin\end{tabular} \\
6 & \begin{tabular}[c]{@{}c@{}}ic\_1, bc\_2\end{tabular} & \begin{tabular}[c]{@{}c@{}}ic\_1, bc\_2\end{tabular} &  & pde\_1 &  &  \\
7 & bc\_2 & bc\_2,pde\_1 & ic\_1 & ic\_1 &  &  \\
8 & ic\_1 & ic\_1,bc\_4,pde\_1 & bc\_2,bc\_3,bc\_4 & bc\_2,bc\_3 &  & bc\_4 is small \\
9 & bc\_2 & bc\_2,pde\_1 & ic\_1 & ic\_1 &  &  \\
10 & \begin{tabular}[c]{@{}c@{}}bc\_2,bc\_3,\\bc\_4,bc\_5\end{tabular} & \begin{tabular}[c]{@{}c@{}}bc\_2,bc\_3,bc\_4,\\bc\_5,pde\_1,pde\_2,\\pde\_3\end{tabular} & bc\_1 & bc\_1 &  &  \\
11 & \begin{tabular}[c]{@{}c@{}}bc\_2,bc\_3,\\bc\_4,bc\_5\end{tabular} & \begin{tabular}[c]{@{}c@{}}bc\_2,bc\_3,bc\_4,\\bc\_5,pde\_1,pde\_2,\\pde\_3\end{tabular} & bc\_1 & bc\_1 &  &  \\
12 & \begin{tabular}[c]{@{}c@{}}bc\_2,bc\_3,bc\_4,\\bc\_5,bc\_6,bc\_7,\\bc\_8\end{tabular} & \begin{tabular}[c]{@{}c@{}}bc\_2,bc\_3,bc\_4,\\bc\_5,bc\_6,bc\_7,\\bc\_8,pde\_1,pde\_3\end{tabular} & bc\_1 & bc\_1,pde\_2 &  &  \\
13 & bc\_1,bc\_3 & bc\_1,bc\_3,pde\_1 & bc\_2 & bc\_2 &  &  \\
14 & bc\_2, bc\_3 & bc\_2,bc\_3,pde\_1 & bc\_1 & bc\_1 &  &  \\
15 &  &  & ic\_1,bc\_2 & \begin{tabular}[c]{@{}c@{}}ic\_1,bc\_2,\\pde\_1\end{tabular} &  &  \\
16 &  & pde\_1 &  bc\_1 & bc\_1 &  &  \\
17 &  &  & ic\_1,ic\_2 & \begin{tabular}[c]{@{}c@{}}ic\_1,ic\_2\\pde\_1,pde\_2\end{tabular} &  &  \\
18 &  &  & bc\_1 & bc\_1,pde\_1 &  &  \\
19 &  & pde\_1  &bc\_1,ic\_2 & bc\_1,ic\_2&  &  \\
\bottomrule
\end{tabular}
\end{adjustbox}
\vspace{0.2cm}
\raggedright
\small
\textit{Note:}\\
Components with zero boundary/initial conditions (BC/IC) typically start with low loss values and increase during later optimization to balance the total loss, while non-zero BC/IC components start with high loss and decrease over epochs. Cases with ambiguous trends show oscillatory or unclear optimization behavior. Special conditions (periodic, Robin, or small non-zero values) are noted in the Remarks column.

\end{table}
\FloatBarrier
\sect{Foundational Performance on Function Approximation}

To understand the fundamental behavioral differences between our proposed SEA-PINN architecture and a standard Feedforward Neural Network (FNN), we tested them on a simple 1D function approximation task. Both networks were trained to fit four different functions using data only from the domain [0, 1] (the gray shaded region). We evaluated their performance on a wider domain from [-0.5, 1.5] to observe their behavior outside the training data (the white regions). The experiment was repeated 10 times with different random initializations to ensure the results were consistent.

\begin{figure}[b]
				\begin{centering}
					\includegraphics[width=1\textwidth,keepaspectratio]{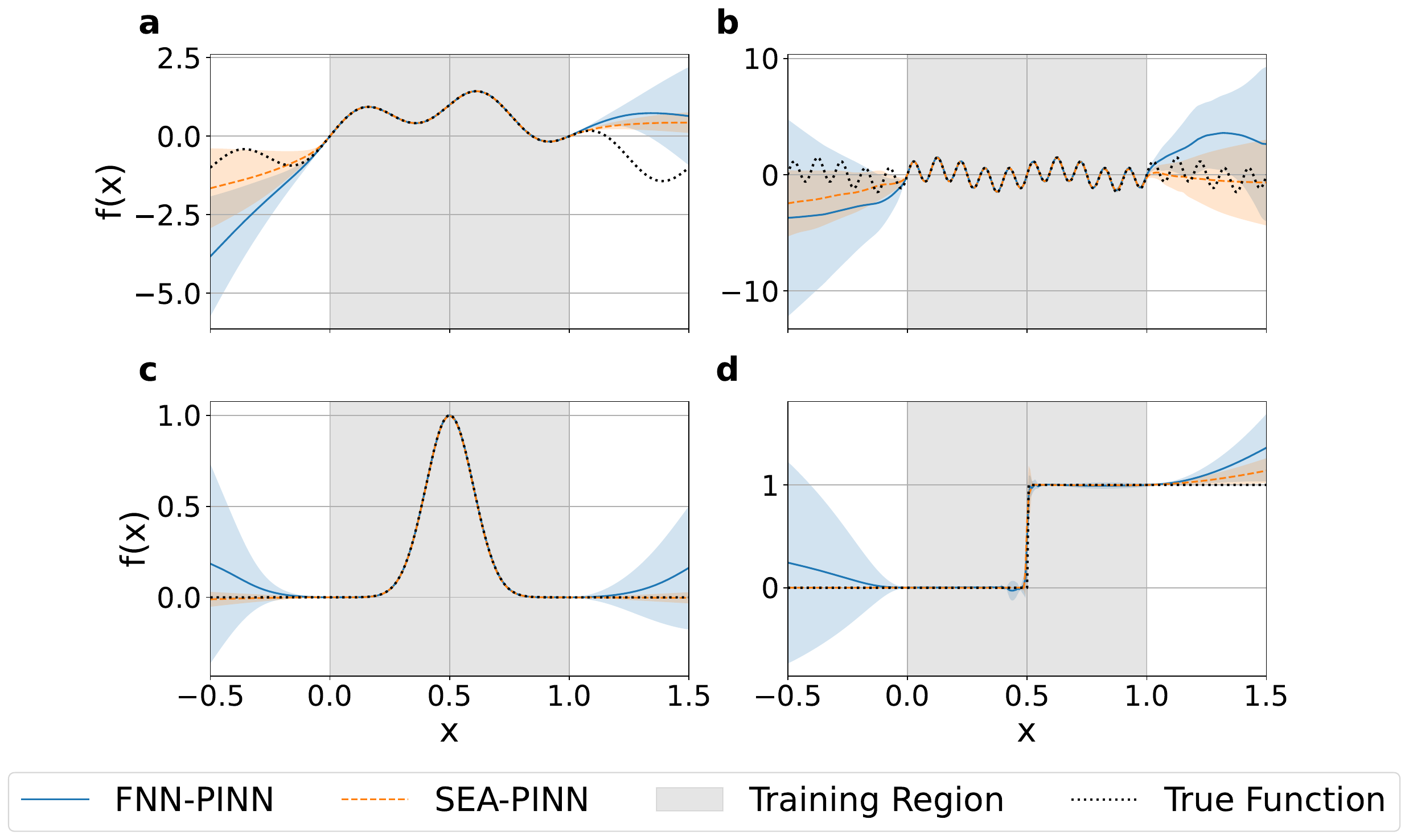}
					\caption{\raggedright 
The panels display the approximation capabilities of FNN-PINN and SEA-PINN across ten independent training runs. The target functions are: \textbf{(a)} a low-frequency sinusoidal function, $f(x) = \sin(\pi x) + 0.5\sin(4\pi x)$; \textbf{(b)} a high-frequency sinusoidal function, $f(x) = \sin(20\pi x) + 0.5\sin(4\pi x)$; \textbf{(c)} a localized SEAussian function, $f(x) = e^{-\frac{(x-0.5)^2}{2\sigma^2}}$  with $\sigma=0.1$; and \textbf{(d)} a discontinuous step function, $f(x) = 1$ for $x > 0.5$ and $0$ otherwise. For each model, the solid line represents the mean prediction over all runs, while the shaded area indicates the standard deviation. The models were trained on data within the region [0, 1] (gray shaded area) and tested on the extended domain [-0.5, 1.5]. }
					\label{fig:Function Approximation}
				\end{centering}
			\end{figure}

Fig. \ref{fig:Function Approximation} shows the average prediction (solid lines) and the standard deviation (shaded bands) for both models. The results show two key points:

1. In the top two plots, SEA-PINN's average prediction (orange line) stays closer to the true function (dashed black line) in the extrapolation regions ( x $<$ 0 and x $>$ 1 ). Furthermore, its uncertainty band (orange shading) is significantly smaller than the FNN-PINN's (blue shading), indicating that SEA-PINN's predictions are not only more accurate but also more consistent and reliable when encountering unseen data.
   
2. The bottom two plots, which feature a sharp peak and a sudden jump, highlight this difference even more clearly. While both models learn the difficult features inside the training domain, the FNN-PINN produces highly unstable and incorrect predictions outside of it. In contrast, SEA-PINN defaults to a simple, flat prediction in these unknown regions, maintaining low error and very low uncertainty.

In summary, this experiment demonstrates that SEA-PINN has a more stable and conservative behavior when making predictions outside the domain of seen data. The weight generation mechanism inherent in SEA-PINN naturally suppresses extreme predictions, leading to more physically plausible initial states that facilitate better convergence. This tendency to avoid making wild guesses may be an important factor in its improved performance in the main PINN experiments, as it helps guide the model towards more stable and physically plausible solutions.

\rowcolors{1}{}{}
\normalcolor

\FloatBarrier

\sect{Layer-wise Jacobian norms for representative PDE cases}

\begin{figure}[htbp]
    \centering

    \begin{subfigure}[t]{0.48\textwidth}
        \centering
        \includegraphics[width=\linewidth,keepaspectratio]{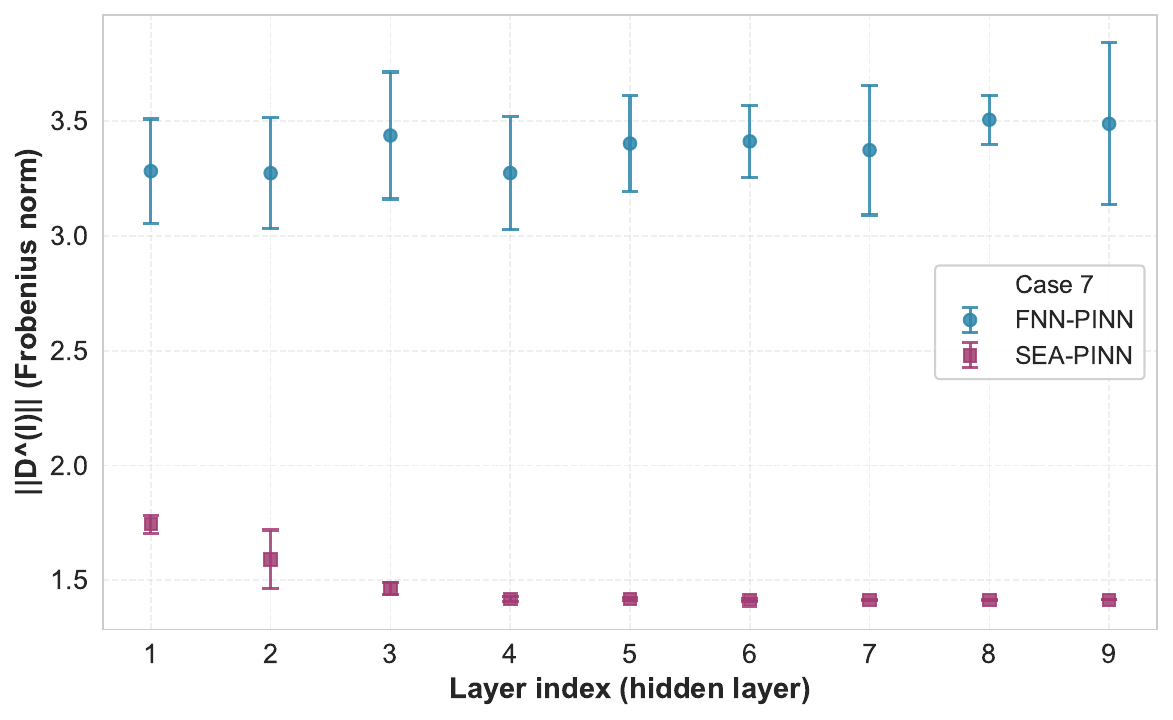}
        \caption{Case~7.}
        \label{fig:case7_D_norms}
    \end{subfigure}
    \hfill
    \begin{subfigure}[t]{0.48\textwidth}
        \centering
        \includegraphics[width=\linewidth,keepaspectratio]{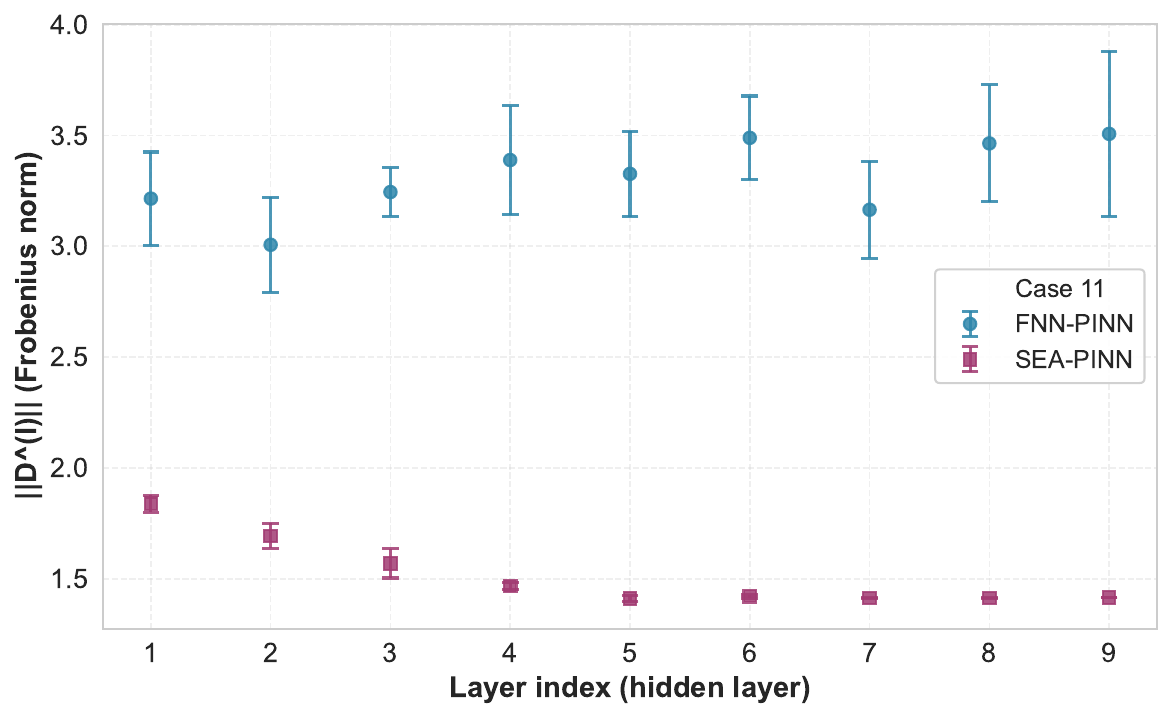}
        \caption{Case~11.}
        \label{fig:case11_D_norms}
    \end{subfigure}
    
    \caption{\raggedright 
    Layer-wise Frobenius norms of local Jacobians for FNN-PINN and SEA-PINN.
    For each hidden layer $l$ in two representative PDE problems (Cases~7 and~11), we evaluate the Frobenius norm of the local Jacobian $\|\mathbf{D}^{(l)}\|_F = \bigl\|\partial \mathbf{h}^{(l)} / \partial \mathbf{z}^{(l)}\bigr\|_F$ at initialization, for both FNN-PINN and SEA-PINN.
    In each case, three random seeds are used; for every seed, the norm at each layer is averaged over $1,000$ randomly selected training points, and error bars indicate the standard deviation across seeds.
    Across all hidden layers and in both cases, SEA-PINN exhibits consistently smaller layer-wise Jacobian norms and reduced variability than FNN-PINN, supporting the interpretation that SEA-PINN realizes smoother and more stable feature transformations throughout the network.
    }
    \label{fig:per_layer_D_norms}

\end{figure}
\FloatBarrier
\sect{Heatmap}

Case10 (NS2D\_LidDriven) models the lid-driven flow within a 2D square cavity. The problem is governed by the steady-state incompressible Navier-Stokes equations, identical to Case 11 in the main text, and similarly constitutes a steady-state, multi-output problem. The computational domain is a unit square $[0, 1]\times [0,1]$.

The key distinction from Case 11 lies in the boundary conditions: the top wall ($y=1$) moves with a parabolic horizontal velocity profile, which drives the flow. No-slip conditions ($u=0$, $v=0$) are applied on the other three walls. The pressure is fixed at a reference value of 0 at the point $(0, 0)$. This setup generates characteristic recirculation vortices in the cavity corners, making it a classic benchmark for validating numerical methods in handling fluid motion within enclosed domains.

\begin{figure}[htbp]
\centering

    \includegraphics[width=0.83\linewidth,keepaspectratio]{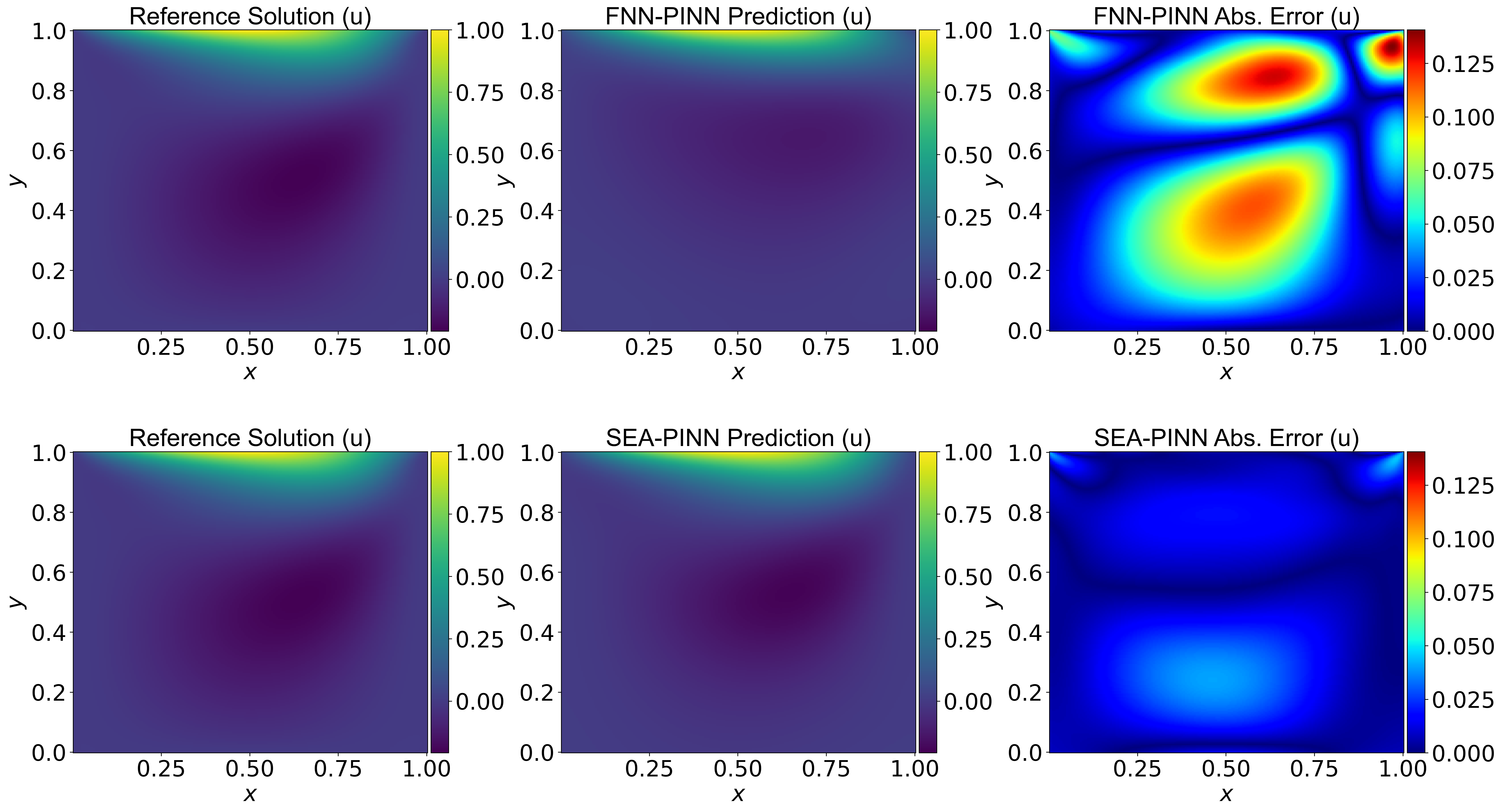}\\

    \includegraphics[width=0.83\linewidth,keepaspectratio]{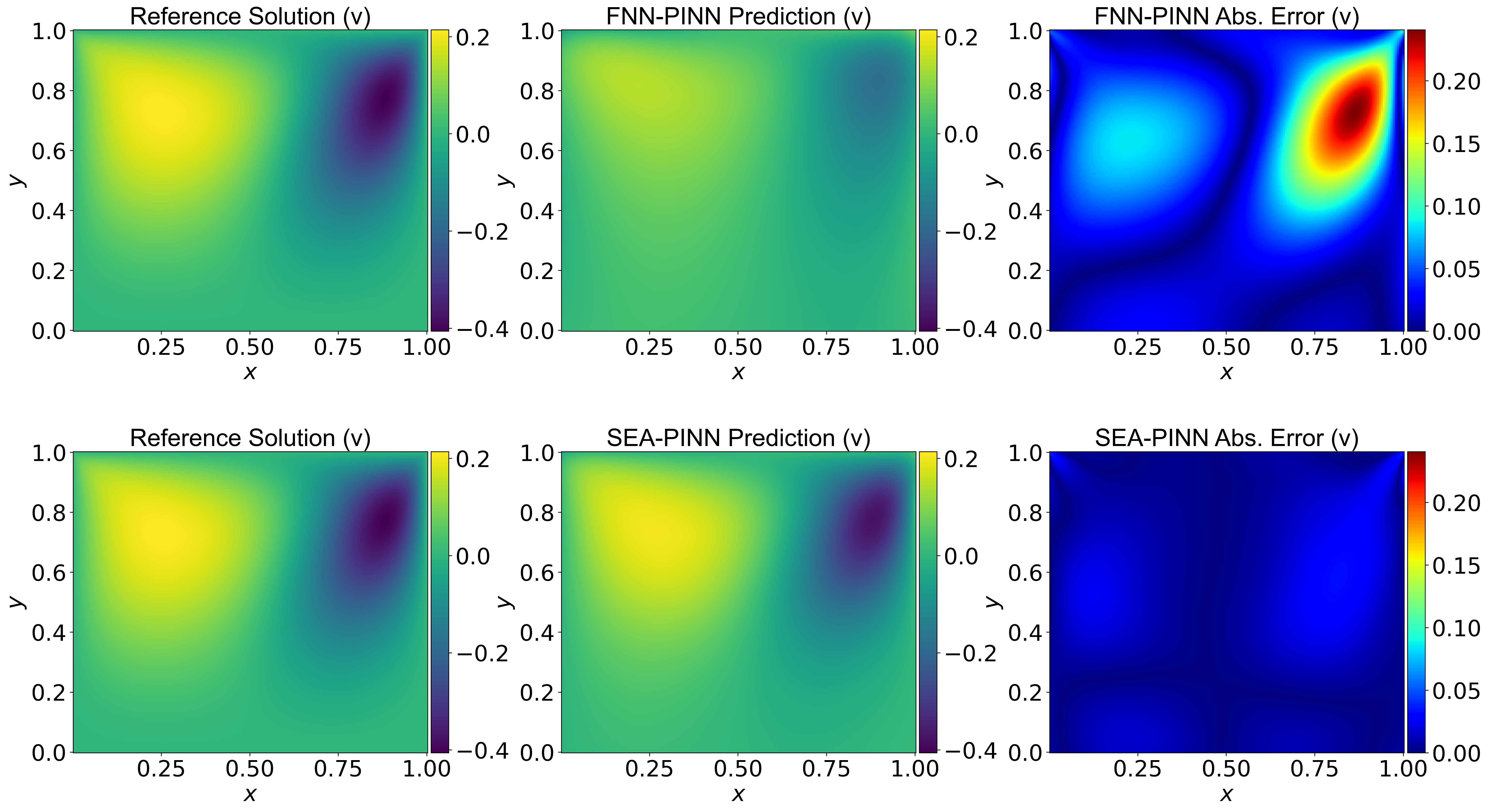}\\

    \includegraphics[width=0.83\linewidth,keepaspectratio]{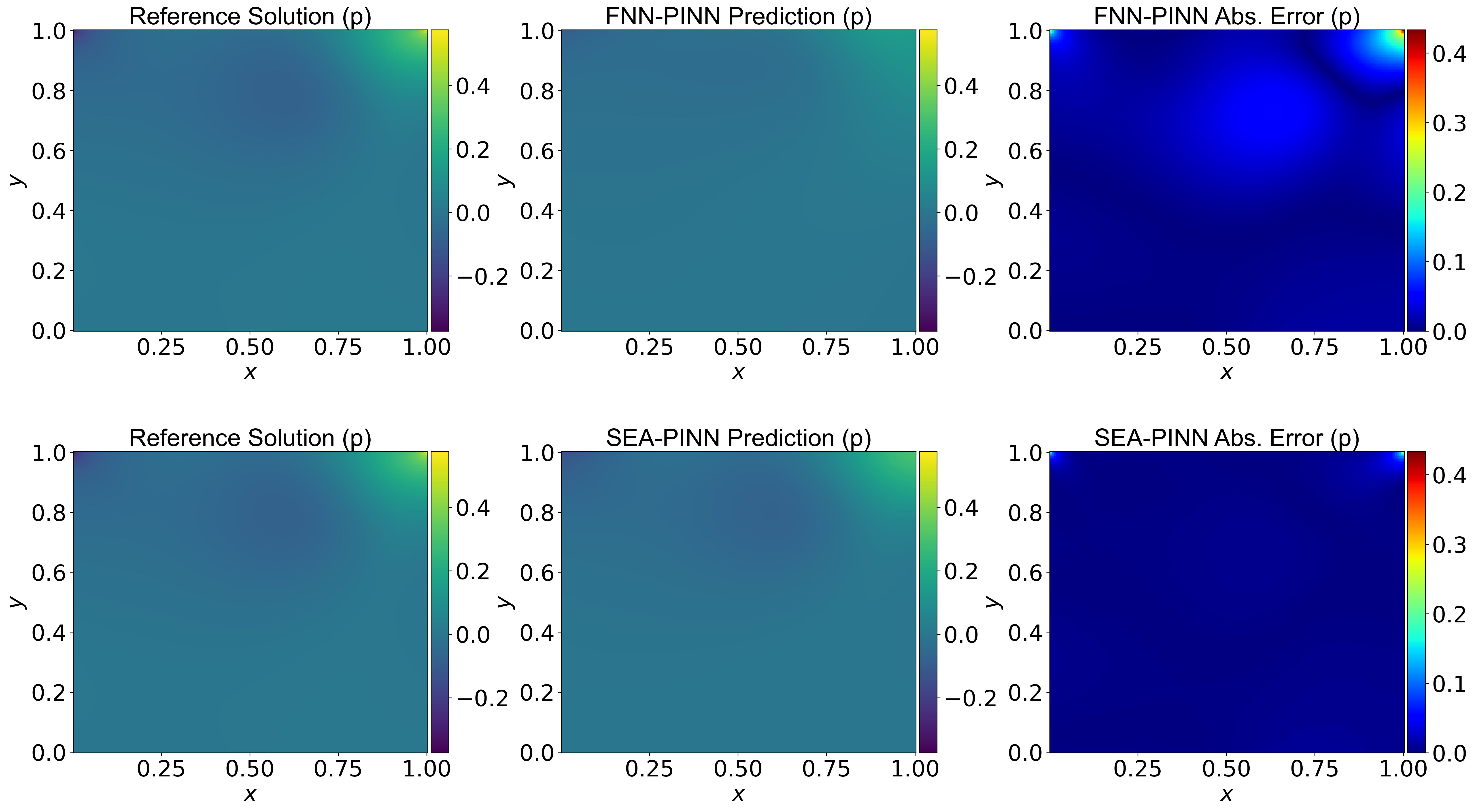}\\

\caption
  {\raggedright 
  NS2D\_LidDriven(ID=10):Heatmaps of FNN-PINN and SEA-PINN at Different Components(Seed=1),
  Top: $x$-velocity, 
  Middle: $y$-velocity, 
  Bottom: pressure }
\label{fig:heatmap_case10}
\end{figure}

\begin{figure}[htbp]
\centering

    \includegraphics[width=0.83\linewidth,keepaspectratio]{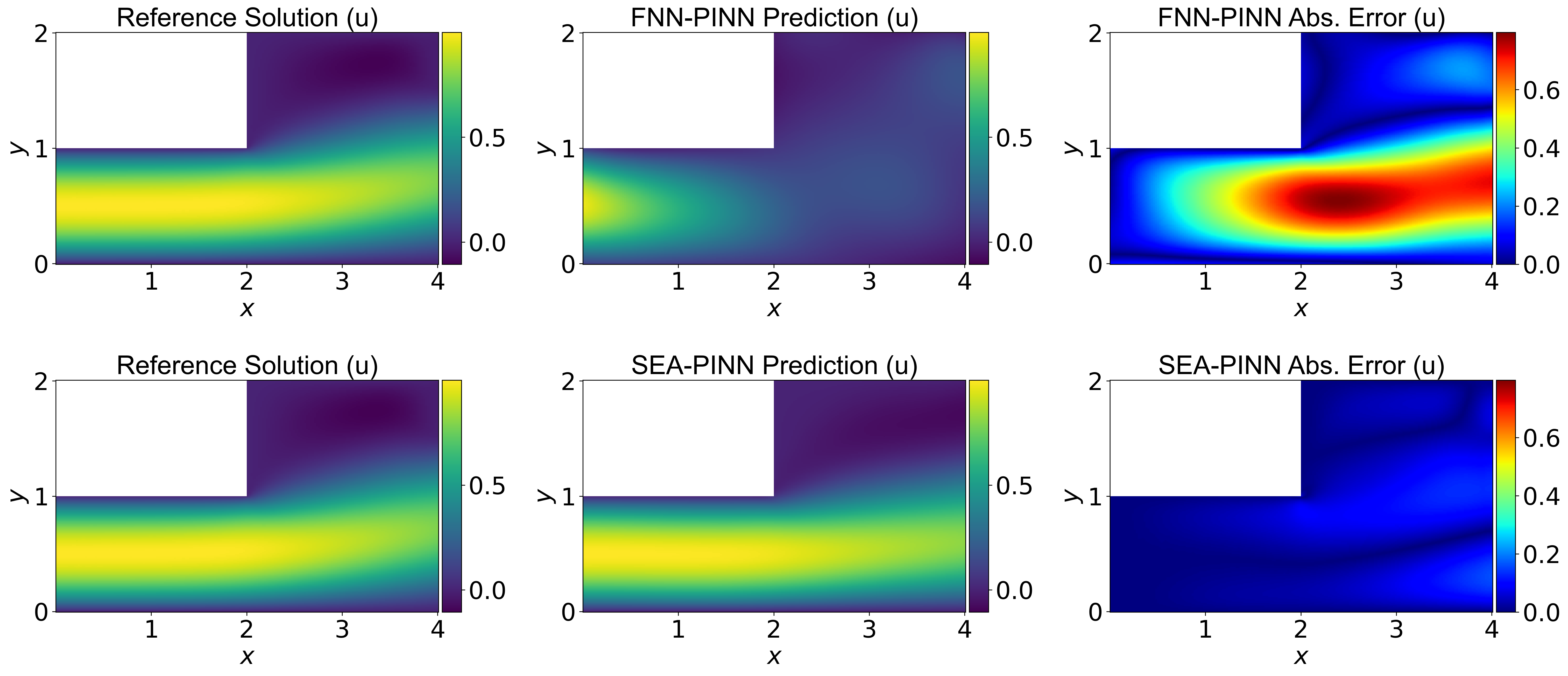}\\

    \includegraphics[width=0.83\linewidth,keepaspectratio]{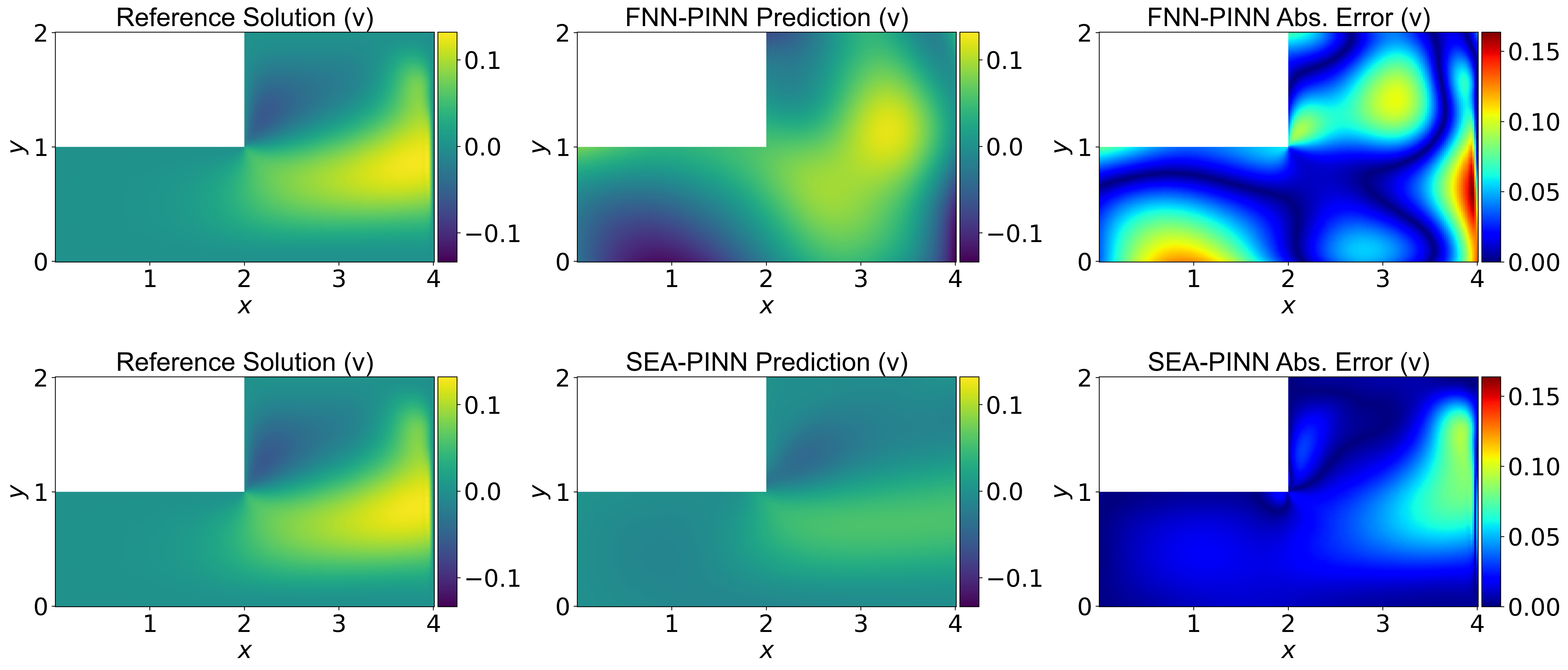}\\

    \includegraphics[width=0.83\linewidth,keepaspectratio]{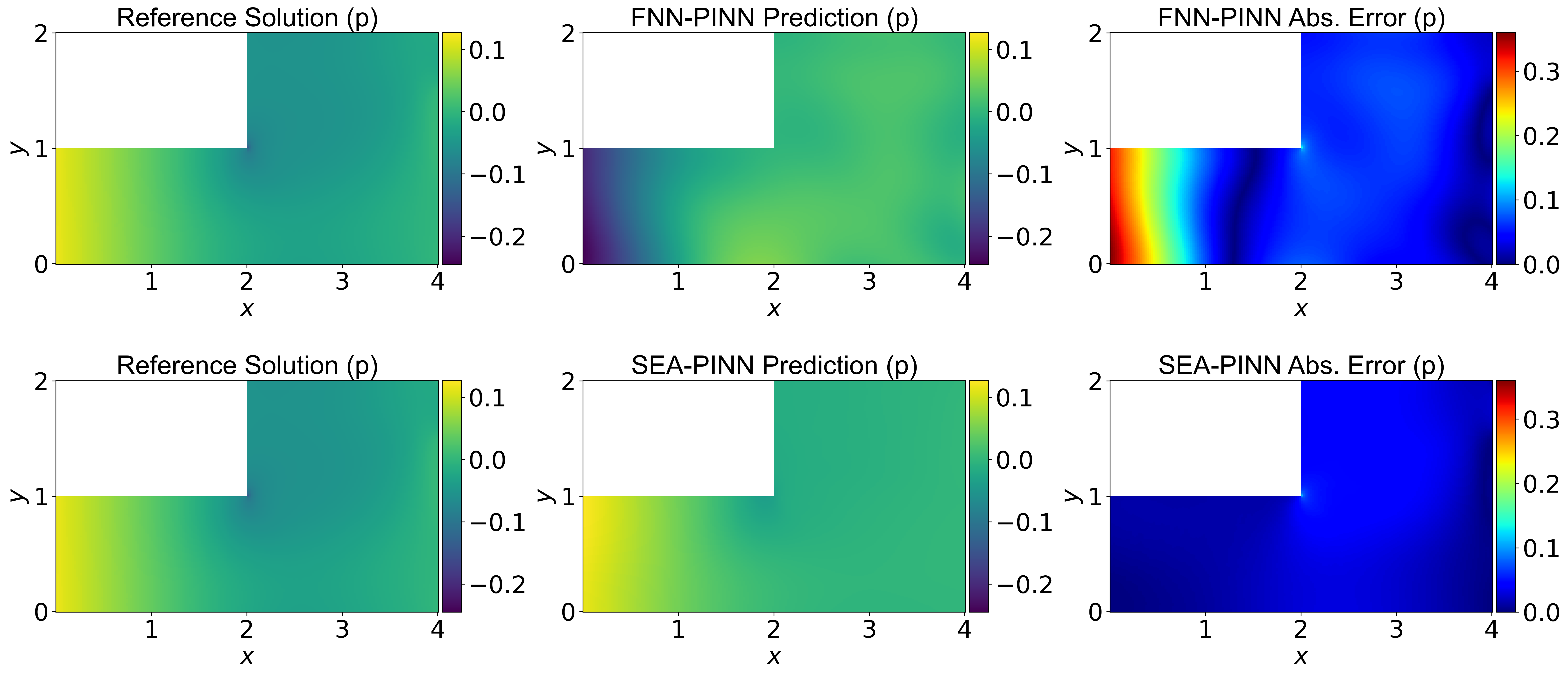}\\

\caption
  {\raggedright 
  NS2D\_BackStep(ID=11):Heatmaps of FNN-PINN and SEA-PINN at Different Components(Seed=1),
  Top: $x$-velocity, 
  Middle: $y$-velocity, 
  Bottom: pressure }
\label{fig:heatmap_case11}
\end{figure}

Fig. \ref{fig:heatmap_case10} and Fig. \ref{fig:heatmap_case11} compares the output solutions from SEA-PINN and FNN-PINN for Cases 10 and 11 using heatmaps.

\FloatBarrier

\sect{Comparison of Relative $L^2$ Errors Across Different Architectures}

\begin{figure}[!htbp] 
\centering
\includegraphics[width=\linewidth,keepaspectratio]{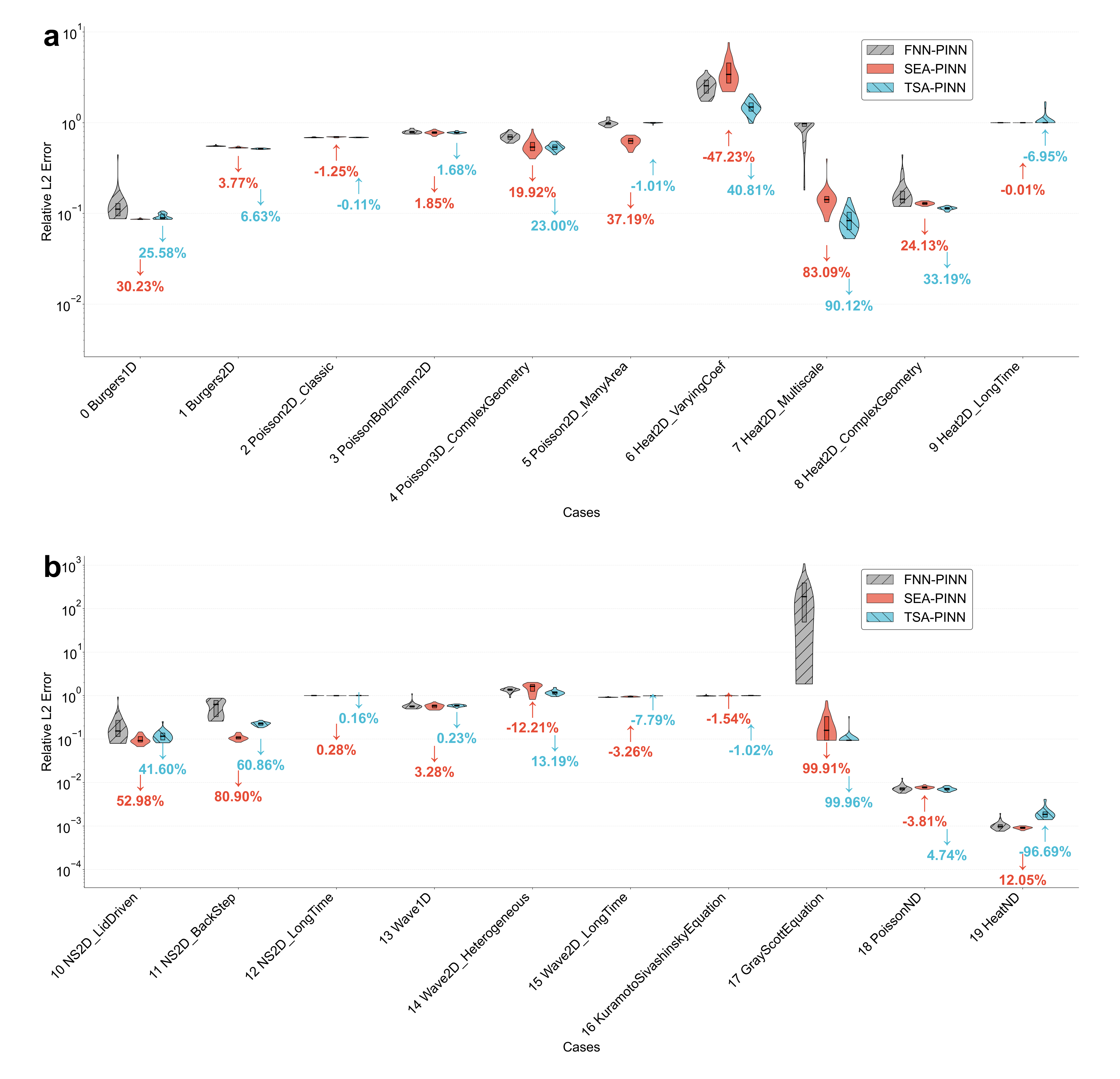} 
\caption{\raggedright 
Performance comparison of different architectures. 
        Comparison of relative $L^{2}$ errors for FNN-PINN, SEA-PINN, and TSA-PINN across 20 PDE cases (30 seeds/case). Percentage labels show improvement over FNN-PINN baseline.The colored percentage labels, along with their corresponding arrows, quantify the performance change of SEA-PINN and TSA-PINN relative to the FNN-PINN baseline. A downward arrow indicates an improvement, whereas an upward arrow indicates a degradation.  }
\label{fig:violin_cases_0-19_combined}

\end{figure}
`

\begin{table}[!htbp]

    \rowcolors{1}{}{}
    \normalcolor
    
\begingroup 
\centering

\caption{Percentage Improvement of Various Architectures Over the FNN-PINN}
\label{tab:pinn_improvement}
\begin{threeparttable} 
\setlength{\tabcolsep}{4pt}
\begin{tabular}{@{}llccc@{}}

\hline 
\textbf{Case ID} & \textbf{Case Name} & \multicolumn{1}{c}{\textbf{SEA-PINN}} & \multicolumn{1}{c}{\textbf{TSA-PINN}} & \multicolumn{1}{c}{\textbf{TSA\_SEA-PINN}} \\
& & \multicolumn{1}{c}{\textbf{Improvement(\%)}} & \multicolumn{1}{c}{\textbf{Improvement(\%)}} & \multicolumn{1}{c}{\textbf{Improvement(\%)}} \\

\hline 
0 & Burgers1D & 30.23 & 25.58 & \textcolor{red}{30.30} \\
1 & Burgers2D & 3.77 & 6.63 & 5.63 \\
2 & Poisson2D\_Classic & -1.25 & -0.11 & -0.63 \\
3 & PoissonBoltzmann2D & 1.85 & 1.68 & \textcolor{red}{1.86} \\
4 & Poisson3D\_Complex & 19.92 & 23.00 & \textcolor{red}{27.00} \\
5 & Poisson2D\_ManyArea & 37.19 & -1.01 & \textcolor{red}{54.34} \\
6 & Heat2D\_VaryingCoef & -47.23 & 40.81 & 34.57 \\
7 & Heat2D\_Multiscale & 83.09 & 90.12 & \textcolor{red}{94.32} \\
8 & Heat2D\_Complex & 24.13 & 33.19 & \textcolor{red}{35.11} \\
9 & Heat2D\_LongTime & -0.01 & -6.95 & -0.36 \\
10 & NS2D\_LidDriven & 52.98 & 41.60 & \textcolor{red}{74.04} \\
11 & NS2D\_BackStep & 80.90 & 60.86 & \textcolor{red}{82.08} \\
12 & NS2D\_LongTime & 0.28 & 0.16 & 0.15 \\
13 & Wave1D & 3.28 & 0.23 & -54.36 \\
14 & Wave2D\_Hetero & -12.21 & 13.19 & \textcolor{red}{34.32} \\
15 & Wave2D\_LongTime & -3.26 & -7.79 & -8.66 \\
16 & KuramotoSivashinsky & -1.54& -1.02 & -0.53 \\
17 & GrayScottEquation & 99.91 & 99.96 & 99.96 \\
18 & PoissonND & -3.81 & 4.74 & 3.01 \\
19 & HeatND & 12.05 & -96.69 & -63.57 \\
\hline 

\end{tabular}
\begin{tablenotes}
\footnotesize
\item[*] The red data indicate cases where TSA\_SEA-PINN outperforms both SEA-PINN and TSA-PINN
\end{tablenotes}
\end{threeparttable}
\endgroup 
\end{table}

\FloatBarrier

\sect{The Comparison of SEA-PINN Variants}

The network depth of the Weight Generator (WG) can be manually adjusted to improve accuracy by simply increasing or decreasing the number of its layers. The SEA-PINN architecture used in the main text is denoted as the Standard SEA-PINN. By slightly increasing or decreasing the number of layers in the WG, we obtain two variants: SEA-PINN-LargeWG and SEA-PINN-SmallWG. 

\begin{itemize}
    \item SEA-PINN-LargeWG: Increases the weight generator’s complexity to three hidden layers, structured as (linear layer + Tanh) × 3 + linear layer + Sigmoid.
    \item SEA-PINN-SmallWG: Simplifies the weight generator to a single linear layer, structured as linear layer + Sigmoid.
\end{itemize}

These two variants are trained and tested under the same settings as the Standard SEA-PINN. 

\begin{figure}
    \rowcolors{1}{}{}
    \normalcolor
    
\centering
\begin{tabular}{c}
\includegraphics[width=\linewidth,keepaspectratio]{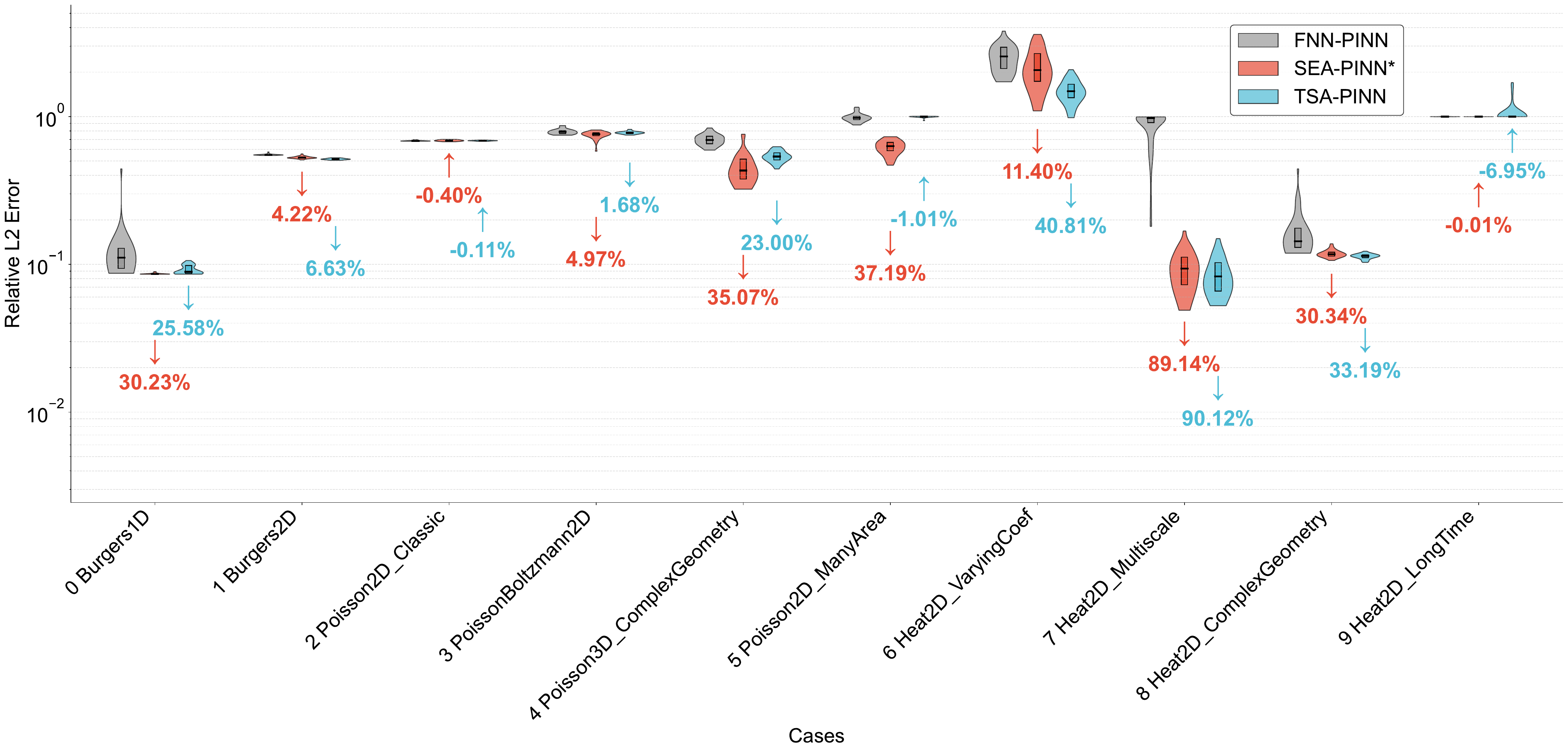} \\

\includegraphics[width=\linewidth,keepaspectratio]{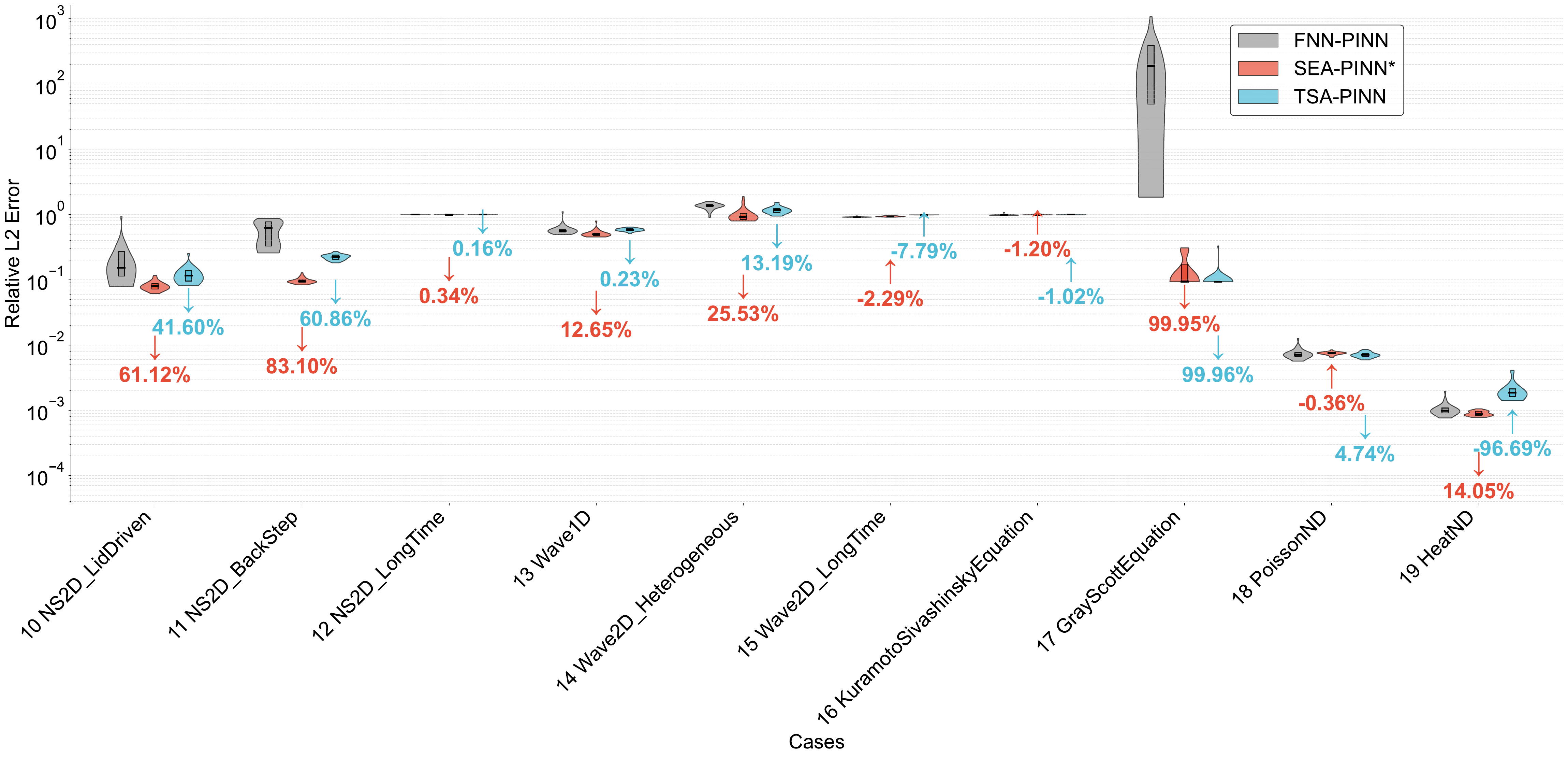} \\
\end{tabular}

\caption{\raggedright 
Performance comparison of different architectures. 
Comparison of relative $L^{2}$ errors for FNN-PINN, SEA-PINN*, and TSA-PINN across 20 PDE cases (30 seeds/case). Percentage labels show improvement over FNN-PINN baseline.The colored percentage labels, along with their corresponding arrows, quantify the performance change of SEA-PINN (orange) and TSA-PINN (cyan) relative to the FNN-PINN baseline.A downward arrow indicates a reduction in the relative $L^2$ error, signifying an improvement in performance, while an upward arrow indicates an increase in error, corresponding to a degradation in performance. SEA-PINN* is defined as the optimal variant architecture selected from Standard SEA-PINN, SEA-PINN-LargeWG, and SEA-PINN-SmallWG for each individual case.  }
\label{fig:SEA-PINN-star}

\end{figure}

Table \ref{tab:SEA-PINN_variants_comparison} presents the test results obtained from 30 experimental runs each for the Standard SEA-PINN and its two variants, all trained on 20 distinct cases under identical settings. By increasing or decreasing the number of WG layers, higher accuracy was achieved in 17 out of the 20 cases. It is noteworthy that the optimal number of WG layers depends on the specific problem; a WG with more layers does not always guarantee better performance. In certain scenarios, a simpler WG may yield superior results. For each case, the network with the most suitable WG is denoted as SEA-PINN. As shown in Fig. \ref{fig:SEA-PINN-star}, we compare SEA-PINN against FNN-PINN and TSA-PINN over 30 random experiments. SEA-PINN* demonstrates significant improvement across a wide range of cases, with smaller fluctuations in accuracy over multiple runs compared to FNN-PINN and TSA-PINN. Therefore, by manually adjusting the depth of the WG network, we can further enhance the accuracy and stability of our model.

    \begin{table}[htbp]
    \rowcolors{1}{}{}
    \normalcolor
    
        \centering
        \caption{Performance Comparison of Standard SEA-PINN and Variants Across PDE Cases}
        \label{tab:SEA-PINN_variants_comparison}
        \begin{adjustbox}{max width=\textwidth} 
        \begin{threeparttable} 
        \setlength{\tabcolsep}{4pt}
        \begin{tabular}{c l S[table-format=1.4e-1] S[table-format=1.4e-1] S[table-format=1.4e-1]} 
            \hline 
            \textbf{Case ID} & \textbf{Case Name} & \multicolumn{1}{c}{\textbf{Standard SEA-PINN}} & \multicolumn{1}{c}{\textbf{SEA-PINN-LargeWG}} & \multicolumn{1}{c}{\textbf{SEA-PINN-SmallWG}} \\ 
            & & {Rel. $L^2$ Error} & {Rel. $L^2$ Error} &  {Rel. $L^2$ Error} \\ 
            \hline 
            0 & Burgers1D & \textcolor{red}{\textbf{8.6339e-02}} & 8.6942e-02 & 8.7045e-02 \\
            1 & Burgers2D & 5.3134e-01 & 5.2960e-01 & \textcolor{red}{\textbf{5.2883e-01}} \\
            2 & Poisson2D\_Classic & 6.9451e-01 & \textcolor{red}{\textbf{6.8871e-01}} & 6.9457e-01 \\
            3 & PoissonBoltzmann2D & 7.7655e-01 & 8.0679e-01 & \textcolor{red}{\textbf{7.5187e-01}} \\
            4 & Poisson3D\_ComplexGeometry & 5.6022e-01 & \textcolor{red}{\textbf{4.5427e-01}} & 5.6941e-01 \\
            5 & Poisson2D\_ManyArea & \textcolor{red}{\textbf{6.1986e-01}} & 6.6602e-01 & 7.0514e-01 \\
            6 & Heat2D\_VaryingCoef & 3.7379e+00 & 5.2684e+00 & \textcolor{red}{\textbf{2.2493e+00}} \\
            7 & Heat2D\_Multiscale & 1.4949e-01 & 3.0402e-01 & \textcolor{red}{\textbf{9.5947e-02}} \\
            8 & Heat2D\_ComplexGeometry & 1.2882e-01 & 1.3003e-01 & \textcolor{red}{\textbf{1.1829e-01}} \\
            9 & Heat2D\_LongTime & \textcolor{red}{\textbf{9.9922e-01}} & 9.9923e-01 & 1.0478e+00 \\
            10 & NS2D\_LidDriven & 9.8668e-02 & 9.8498e-02 & \textcolor{red}{\textbf{8.1573e-02}} \\
            11 & NS2D\_BackStep & 1.0921e-01 & 1.1041e-01 & \textcolor{red}{\textbf{9.6624e-02}} \\
            12 & NS2D\_LongTime & 9.9711e-01 & 9.9719e-01 & \textcolor{red}{\textbf{9.9651e-01}} \\
            13 & Wave1D & 5.6748e-01 & 6.5171e-01 & \textcolor{red}{\textbf{5.1253e-01}} \\
            14 & Wave2D\_Heterogeneous & 1.5180e+00 & \textcolor{red}{\textbf{1.0074e+00}} & 1.5058e+00 \\
            15 & Wave2D\_LongTime & 9.4565e-01 & \textcolor{red}{\textbf{9.3676e-01}} & 9.6559e-01 \\
            16 & KuramotoSivashinskyEquation & 1.0009e+00 & \textcolor{red}{\textbf{9.9756e-01}} & 1.0146e+00 \\
            17 & GrayScottEquation & 2.3124e-01 & \textcolor{red}{\textbf{1.4238e-01}} & 4.6128e+00 \\
            18 & PoissonND & 7.7097e-03 & \textcolor{red}{\textbf{7.4540e-03}} & 7.7846e-03 \\
            19 & HeatND & 9.1405e-04 & 9.0488e-04 & \textcolor{red}{\textbf{8.9328e-04}} \\
            \hline 
        \end{tabular}
        \begin{tablenotes}
\footnotesize
\item[*] The red data indicate the lowest relative $L^2$ error among the three networks.
\end{tablenotes}
\end{threeparttable}
\end{adjustbox}
    \end{table}

\FloatBarrier

\sect{Extended Experimental Evaluation}

For a more comprehensive comparison, we trained SEA-PINN alongside other network architectures, including GAAF, LAAF, and the baseline FNN-PINN, for 20,000 epochs. In this extended benchmark,
we adopt a different architecture: fully connected networks with 5 hidden layers, 100 neurons per hidden layer, \texttt{tanh} activations, and Xavier-normal initialization. The relative L2 error was evaluated on the test set every 100 epochs, with comparisons based on the best-performing results. Each method was run three times with different random seeds, and the average values are reported. To ensure statistical reliability, we maintained the same number of collocation points as described in the main text for 2D input cases, while for 3D input problems, we used four times the number of collocation points. In addition to the 20 forward problems tested in the main text, we included two additional inverse problems (PoissonInv and HeatInv) for comparison. Note that this configuration differs from the SEA-PINN setting used in the main text, but within this section all methods share exactly the same architecture and training protocol, ensuring a fair comparison under this Extended Experimental.

    \rowcolors{1}{}{}
    \normalcolor

\begin{table}[htbp]
\centering
\begin{threeparttable} 
\caption{Comparison of different methods on various PDE problems.}
\label{tab:pde_results}
\begin{tabular}{lcccccc}
\toprule
Case & PDE Problem & FNN-PINN & LAAF & GAAF & SEA-PINN \\
\midrule
0 & Burgers1D & 0.0133 & 0.0352 & 0.6539 & \cellcolor{mincolor}0.0127 \\
1 & Burgers2D & 0.5175 & 0.5196 & 0.5377 & \cellcolor{mincolor}0.4941 \\
2 & Poisson2D\_Classic & 0.6735 & \cellcolor{mincolor}0.6287 & 0.6331 & 0.6546 \\
3 & PoissonBoltzmann2D & \cellcolor{mincolor}0.6340 & 0.6384 & 0.9472 & 0.7175 \\
4 & Poisson3D\_ComplexGeometry & 0.2113 & 0.3296 & 0.9449 & \cellcolor{mincolor}0.2031 \\
5 & Poisson2D\_ManyArea & \cellcolor{mincolor}0.6179 & 0.9608 & 1.0020 & 0.7205 \\
6 & Heat2D\_VaryingCoef & 0.9861 & 1.6852 & 0.9993 & 1.0781 \\
7 & Heat2D\_Multiscale & 0.0261 & 0.2138 & 0.9999 & \cellcolor{mincolor}0.0207 \\
8 & Heat2D\_ComplexGeometry & \cellcolor{mincolor}0.0217 & 0.3537 & 0.5829 & 0.1124 \\
9 & Heat2D\_LongTime & 0.9977 & 1.0018 & 1.3085 & 0.9984 \\
10 & NS2D\_LidDriven & 0.0441 & 0.1162 & 1.3443 & \cellcolor{mincolor}0.0373 \\
11 & NS2D\_BackStep & 0.1140 & 0.2346 & 0.8335 & \cellcolor{mincolor}0.1083 \\
12 & NS2D\_LongTime & 0.9944 & 0.9982 & 1.0000 & 0.9953 \\
13 & Wave1D & 0.5228 & 0.7897 & 1.1624 & \cellcolor{mincolor}0.4381 \\
14 & Wave2D\_Heterogeneous & 0.8004 & 0.8087 & \cellcolor{mincolor}0.7821 & 0.7995 \\
15 & Wave2D\_LongTime & 1.0035 & \cellcolor{mincolor}1.0000 & 1.0002 & 1.0031 \\
16 & KuramotoSivashinskyEquation & 0.9702 & 0.9802 & 15.6750 & 0.9719 \\
17 & GrayScottEquation & \cellcolor{mincolor}0.0931 & 0.2469 & \cellcolor{mincolor}0.0931 & \cellcolor{mincolor}0.0931 \\
18 & PoissonND & \cellcolor{mincolor}0.0008 & 0.0121 & 0.1954 & 0.0022 \\
19 & HeatND & 0.0009 & 0.0033 & 0.3530 & \cellcolor{mincolor}0.0003 \\
20 & PoissonInv & 0.0714 & 0.8832 & 1.0297 & \cellcolor{mincolor}0.0612 \\
21 & HeatInv & 1.0969 & 0.3589 & 1.0396 & \cellcolor{mincolor}0.1327 \\
\bottomrule
\end{tabular}

\vspace{0.2cm}
\raggedright
\small
\textit{Note:}\\
Mean L2RE of different PINN architectures on our benchmark (average of three random experiments). Best results are highlighted in \hl{blue}. We do not bold any result if errors of all methods are about 100\% or above 100\%. 
\end{threeparttable}

\end{table}

\sect{Trainable Sine Activation PINN}
\begin{enumerate}
    \item Trainable Sine Activation (TSA)
    
    where the final output of the $k$-th layer is:
    \begin{align}
    \rowcolors{1}{}{}
    \normalcolor
        \mathbf{a}^{(k)}=\left[\begin{array}{c}a_{1}^{(k)} \\a_{2}^{(k)} \\\vdots \\a_{N}^{(k)}\end{array}\right]=\left[\begin{array}{c}\zeta_{1} \sin \left(f_{1}^{(k)} z_{1}^{(k)}\right)+\zeta_{2} \cos \left(f_{1}^{(k)} z_{1}^{(k)}\right) \\\zeta_{1} \sin \left(f_{2}^{(k)} z_{2}^{(k)}\right)+\zeta_{2} \cos \left(f_{2}^{(k)} z_{2}^{(k)}\right) \\\vdots \\\zeta_{1} \sin \left(f_{N}^{(k)} z_{N}^{(k)}\right)+\zeta_{2} \cos \left(f_{N}^{(k)} z_{N}^{(k)}\right)\end{array}\right] 
        \label{eq:TSAPINN2}
    \end{align}
    
    where:
    \begin{itemize}
        \item $z_i^{(k)}$ is the output of the $i$-th neuron in the $k$-th layer,
        
        \item $f_i^{(k)} \in \mathbf{R}$ is an independent trainable frequency parameter for the $i$-th neuron in the $k$-th layer (default initialization: 1.0),
        
        \item $\zeta_1, \zeta_2$ are fixed or trainable scaling coefficients (default: $\zeta_1 = \zeta_2 = 0.5$),
        
        \item $\mathbf{a}^{k} \in \mathbf{R}^{N_k}$ is the input vector from the $k$-th layer,
        
    \end{itemize}
    
    \item Slope Recovery Mechanism
    
    The slope recovery term $S(a)$ dynamically adjusts the slope of the activation function, maintaining active and effective gradient propagation throughout the network.
    
    \begin{align}
        S(a) = \frac{1}{\frac{1}{L-1} \sum_{k=1}^{L-1} \exp\left(\frac{1}{N_k} \sum_{i=1}^{N_k} f_i^{(k)}\right)}
        \label{eq:TSAPINN3}
    \end{align}
    where:
    \begin{itemize}
        \item  $L$ is the total number of layers,
        
        \item $N_k$ is the number of neurons in the $k$-th layer.
    \end{itemize}
    This slope recovery term is incorporated into the loss function to promote rapid recovery of activation slopes:
    \begin{align}
        L_{PINN} = L_{data} + L_{phys} + \lambda S(a)
        \label{eq:TSAPINN4}
    \end{align}
\end{enumerate}

\sect{Hybrid Architecture: TSA\_SEA-PINN}

To further validate the effectiveness and generalizability of our proposed Neuron-wise Adaptive Weighting (SEA-PINN) module, we construct a hybrid model by integrating SEA-PINN with the advanced TSA-PINN architecture, named TSA\_SEA-PINN.

We introduce the weight generator into the hidden layers of TSA-PINN, applying it after the sine activation function to weight each neuron. Specifically, the weighted activation output $\mathbf{\hat{a}}^{(l)}$ of the $l$-th layer is computed as follows:
\begin{enumerate}
    \item Compute the standard TSA-PINN activation output $\mathbf{a}^{(l)}$:  
\begin{align}
a_i^{(l)} = \zeta_1 \sin(f_i^{(l)} z_i^{(l)}) + \zeta_2 \cos(f_i^{(l)} z_i^{(l)})
\label{eq:TSAPINN-SEA-PINN1}
\end{align}
where $z_i^{(l)} = \mathbf{w}_i^{(l)} \cdot \mathbf{\hat{a}}^{(l-1)} + b_i^{(l)}$.

\item Use the activation output $\mathbf{a}^{(l)}$ as input to the weight generator to compute the weight vector $\mathbf{\lambda}^{(l)}$:  
\begin{align}
 \mathbf{\lambda}^{(l)} = \text{Sigmoid}(\mathbf{W}_2 \sigma(\mathbf{W}_1 \mathbf{a}^{(l)} + \mathbf{b}_1) + \mathbf{b}_2)
\label{eq:TSAPINN-SEA-PINN2}
\end{align}

\item Element-wise multiply the weight vector with the activation output to obtain the final layer output:  
\begin{align}
\mathbf{\hat{a}}^{(l)} = \mathbf{\lambda}^{(l)} \odot \mathbf{a}^{(l)}
\label{eq:TSAPINN-SEA-PINN3}
\end{align}
\end{enumerate}
By combining TSA-PINN with our weighting module, the TSA\_SEA-PINN model explores whether the adaptive weighting mechanism can serve as a general enhancement module, further improving the performance of existing advanced PINN architectures.
\FloatBarrier

\sect{Enhancing TSA-PINN with SEA-PINN: Demonstrating Superior Synergistic Performance}

\begin{figure}[htbp] 
\centering
\includegraphics[width=\linewidth, height=0.6\textheight, keepaspectratio]{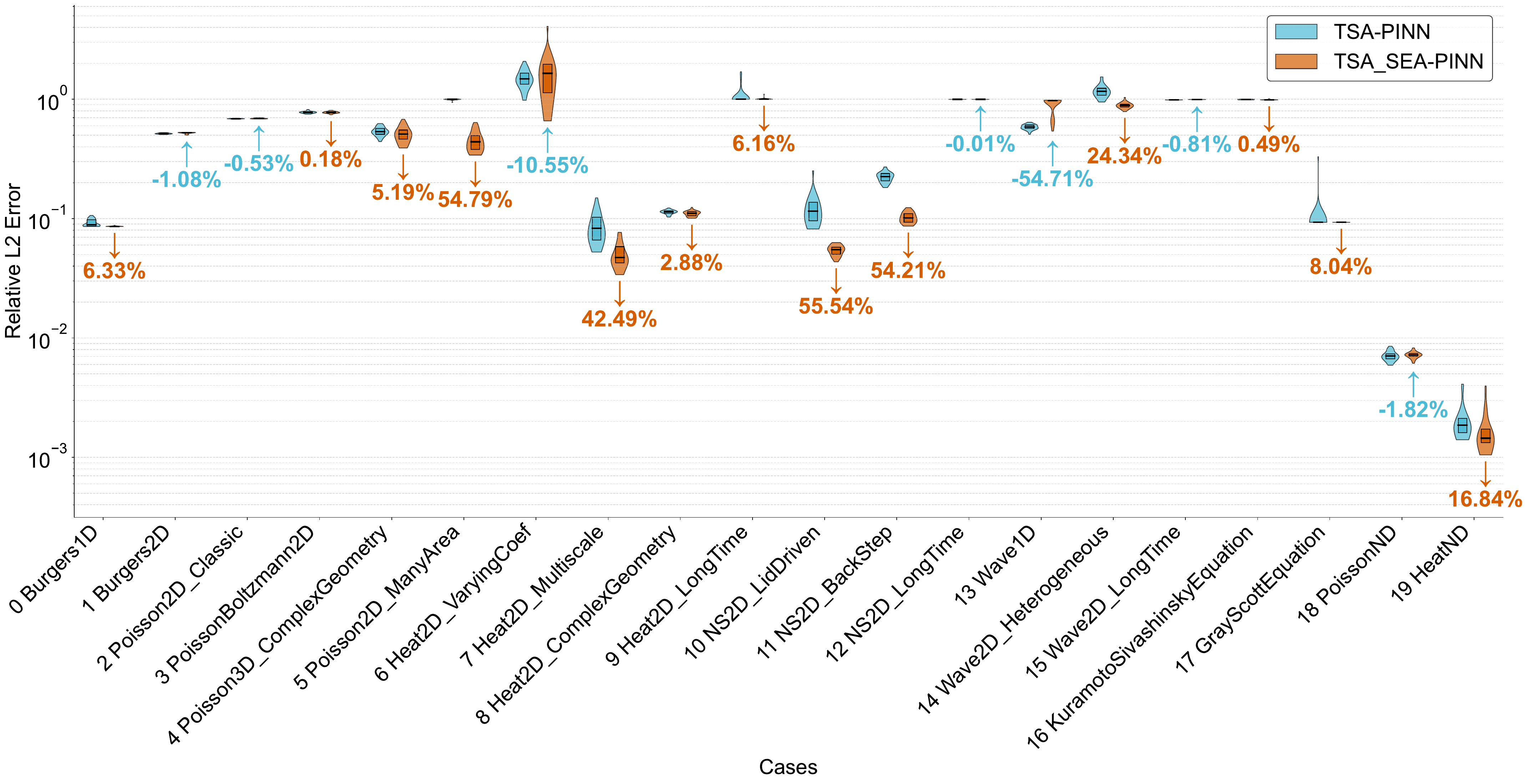}
\caption{\raggedright 
Relative $L^2$ error comparison between the baseline TSA-PINN (blue) and the SEA-PINN-enhanced version (orange) across 20 benchmarks, based on 30 independent runs per case. The percentage indicates the median error reduction achieved by incorporating SEA-PINN. Positive values (orange) signify performance improvement, while negative values (blue) indicate degradation. The results demonstrate a consistent and significant accuracy enhancement from SEA-PINN in most scenarios.}
\label{fig:TSAPINN_SEA-PINN_vs_TSAPINN_comparison}
\end{figure}

The preceding results demonstrate the significant advantages of SEA-PINN over FNN-PINN and its distinct strengths compared to the TSA-PINN architecture. To explore SEA-PINN’s synergistic effects, we integrate it with TSA-PINN to form the hybrid TSA\_SEA-PINN and compare it with the original TSA-PINN. Fig. \ref{fig:TSAPINN_SEA-PINN_vs_TSAPINN_comparison} in Supplementary Information compares the performance of TSA\_SEA-PINN and TSA-PINN across 30 experimental trials, illustrating the relative $L^2$ error improvement percentage of TSA\_SEA-PINN over TSA-PINN. Overall, TSA\_SEA-PINN achieves positive improvements in 13 cases.
For high-frequency problems where TSA-PINN excels, such as Heat2D\_Multiscale (case 7) and GrayScottEquation (case 17), the hybrid architecture TSA\_SEA-PINN achieved improvements of 42.49\% and 8.04\%, respectively, compared to the original model. In problems involving steady-state multi-region and steady-state multi-output scenarios, where our network demonstrates strong performance—such as Poisson2D\_ManyArea (case 5), NS2D\_LidDriven (case 10), and NS2D\_BackStep (case 11)—TSA\_SEA-PINN improved performance by 54.79\%, 55.54\%, and 54.21\%, respectively, consistently showcasing the substantial advantages of our network as a plug-in module.

These results indicate that our network, functioning as a plug-in and collaborating with other advanced architectures, not only enhances performance across a wide range of problems but also exhibits unique strengths in specific types of problems.
\FloatBarrier
\clearpage
\sect{Total Training Loss}
			
			\begin{figure}[!htbp]
				\begin{centering}
					\includegraphics[width=1\textwidth,keepaspectratio]{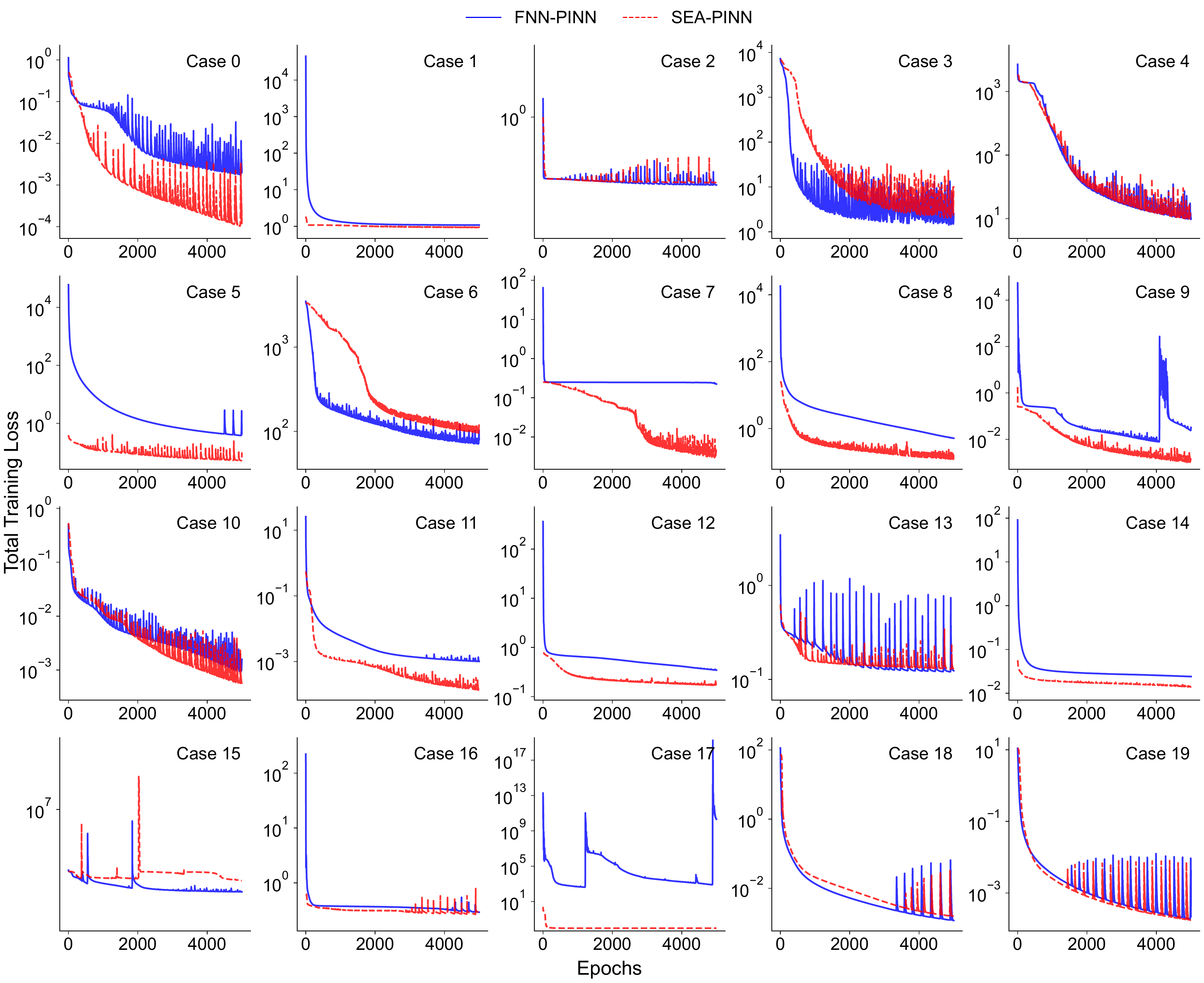}
					\caption{Total Training Loss Comparison: FNN-PINN vs  SEA-PINN (Seed=10) }
					\label{fig:Total training loss_seed10}
				\end{centering}
			\end{figure}

			\begin{figure}[!htbp]
				\begin{centering}
					\includegraphics[width=1\textwidth,keepaspectratio]{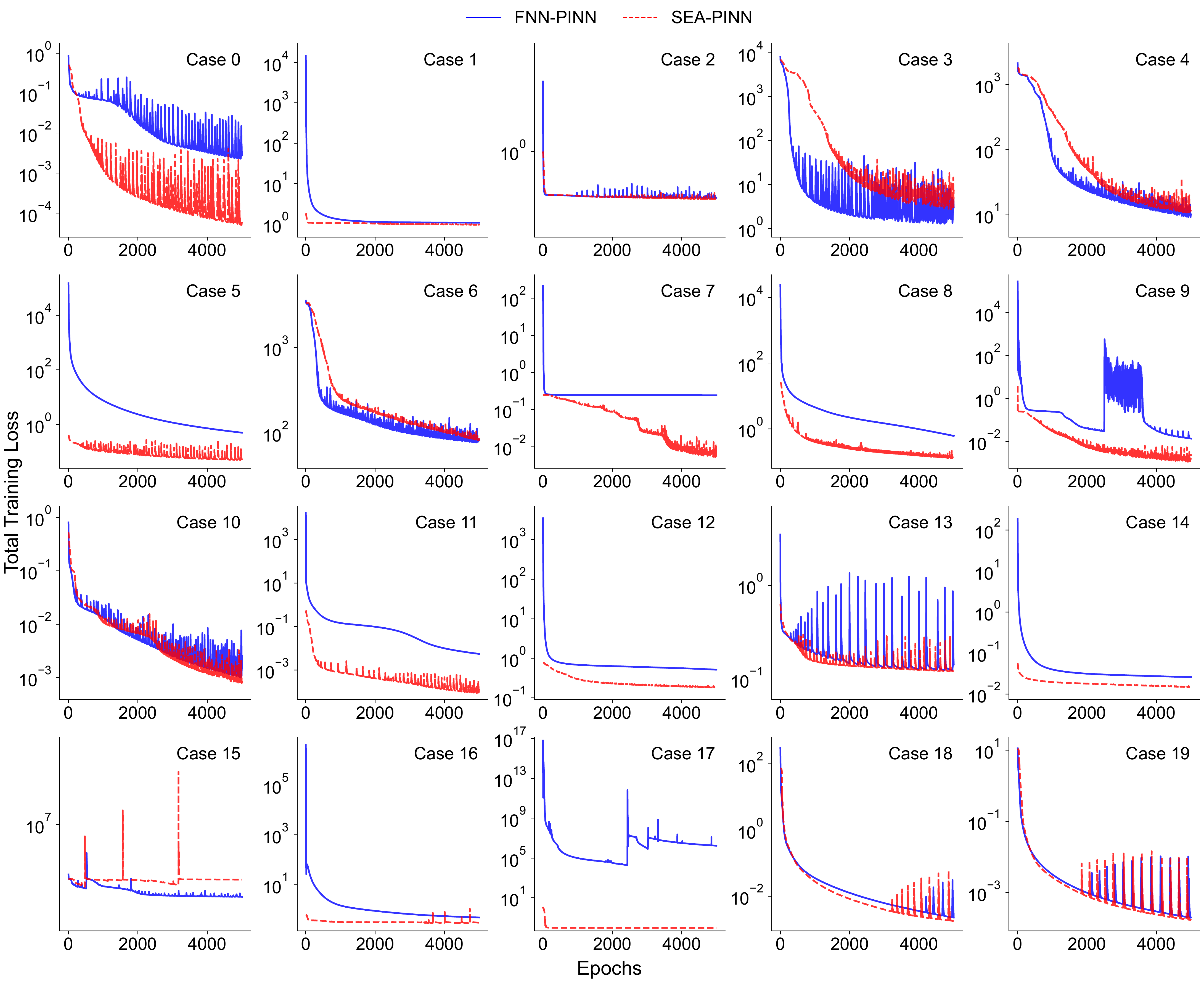}
					\caption{Total Training Loss Comparison: FNN-PINN vs SEA-PINN (Seed=20) }
					\label{fig:Total training loss_seed20}
				\end{centering}
			\end{figure}

			\begin{figure}[!htbp] 
				\begin{centering}
					\includegraphics[width=1\textwidth,keepaspectratio]{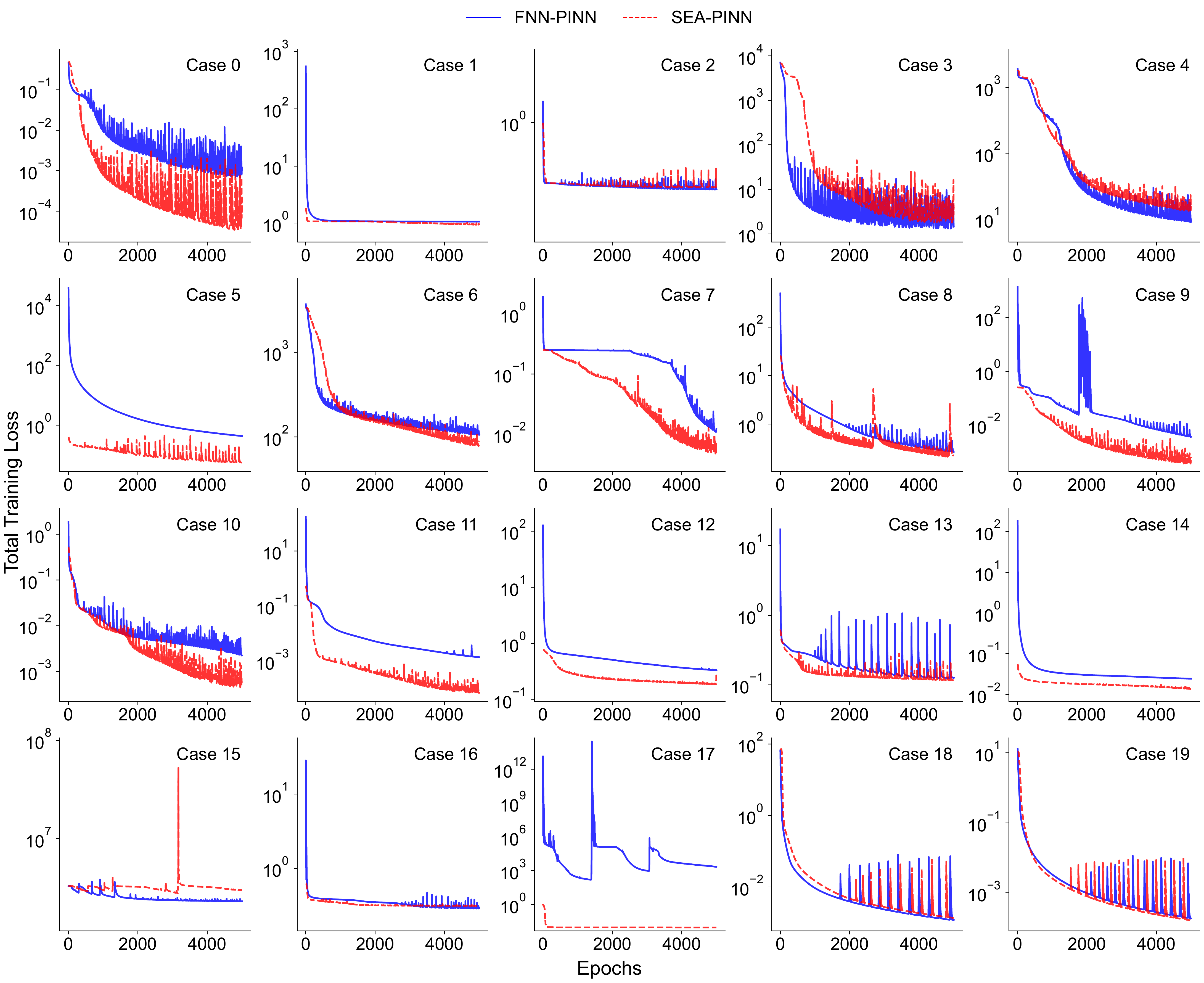}
					\caption{Total Training Loss Comparison: FNN-PINN vs SEA-PINN (Seed=30) }
					\label{fig:Total training loss_seed30}
				\end{centering}
			\end{figure}
		\FloatBarrier

\end{document}